\documentclass[final,3p,times]{elsarticle}
\usepackage{amsmath,amsfonts}
\usepackage{algorithmic}
\usepackage{algorithm}
\usepackage{array}
\usepackage[caption=false,font=normalsize,labelfont=sf,textfont=sf]{subfig}
\usepackage{textcomp}
\usepackage{stfloats}
\usepackage{url}
\usepackage{verbatim}
\usepackage{graphicx}
\usepackage{color}
\usepackage{booktabs}
\usepackage{xcolor}
\usepackage{multirow}
\usepackage{changepage}
\usepackage{soul}
\usepackage{makecell}
\usepackage{longtable}
\soulregister\cite7
\soulregister\ref7

\journal{eClinicalMedicine}

\begin{document}

\begin{frontmatter}

\title{Development and Validation of an AI Foundation Model for Endoscopic Diagnosis of Esophagogastric Junction Adenocarcinoma: a Cohort and Deep Learning Study}

\author[TJIAS]{Yikun Ma$^{\dagger}$}
\author[TJHos]{Bo Li$^{\dagger}$}
\author[TJHos]{Ying Chen$^{\dagger}$}
\author[TJ]{Zijie Yue$^{\dagger}$}
\author[TJHos]{Shuchang Xu}
\author[TJ]{Jingyao Li}
\author[TJ]{Lei Ma}
\author[Huashan]{Liang Zhong}
\author[Ruijin]{Duowu Zou}
\author[Xinhua]{Leiming Xu}
\author[Zhongshan]{Yunshi Zhong}
\author[Renji]{Xiaobo Li}
\author[Huashan]{Weiqun Ding}
\author[Ruijin]{Minmin Zhang}
\author[Zhongshan]{Dongli He}
\author[Xinhua]{Zhenghong Li}
\author[TJHos]{Ye Chen}
\author[Renji]{Ye Zhao}
\author[Huashan]{Jialong Zhuo}
\author[TJHos2]{Xiaofen Wu}
\author[Zhangzhou]{Lisha Yi}
\author[TJ,TJIAS]{Miaojing Shi*}
\author[TJHos]{Huihui Sun*}

\address[TJIAS]{Shanghai Research Institute for Intelligent Autonomous Systems, Tongji University, China.}
\address[TJHos]{Department of Gastroenterology, Tongji Institute of Digestive Disease, Tongji Hospital Affiliated to Tongji University, Medicine of Tongji University, China.}
\address[TJ]{College of Electronic and Information Engineering, Tongji University, China.}
\address[Huashan]{Department of Gastroenterology and Endoscopy, Huashan Hospital Affiliated to Fudan University, China.}
\address[Ruijin]{Department of Gastroenterology, Ruijin Hospital Affiliated to Shanghai Jiao Tong University School of Medicine, China.}
\address[Xinhua]{Department of Gastroenterology, Xinhua Hospital Affiliated to Shanghai Jiao Tong University School of Medicine, China.}
\address[Zhongshan]{Endoscopy Center and Endoscopy Research Institute, Zhongshan Hospital Affiliated to Fudan University, China.}
\address[Renji]{Division of Gastroenterology and Hepatology, Key Laboratory Gastroenterology and Hepatology, Renji Hospital affiliated to Shanghai Jiao Tong University School of Medicine, China.}
\address[TJHos2]{Department of Information Center, Tongji Hospital Affiliated to Tongji University, China.}
\address[Zhangzhou]{Department of Gastroenterology, Zhangzhou Affiliated Hospital of Fujian Medical University, China.}

\footnotetext[1]{$\dagger$Co-first authors.}
\footnotetext[2]{*Corresponding authors.}

\begin{abstract}
\indent\textbf{Background:} The early detection of esophagogastric junction adenocarcinoma (EGJA) is crucial for improving patient prognosis, yet its current diagnosis is highly operator-dependent. This paper aims to make the first attempt to develop an artificial intelligence (AI) foundation model-based method for both screening and staging diagnosis of EGJA using endoscopic images. 

\textbf{Methods:} In this cohort and learning study, we conducted a multicentre study across seven Chinese hospitals between December 28, 2016 and December 30, 2024. It comprises 12,302 images from 1,546 patients (590 with advanced EGJA, 243 with early EGJA, 713 without EGJA); 8,249 of them were employed for model training, while the remaining were divided into the held-out (112 patients, 914 images), external (230 patients, 1,539 images), and prospective (198 patients, 1,600 images) test sets for evaluation. The proposed model employs DINOv2 (a vision foundation model) and ResNet50 (a convolutional neural network) to extract features of global appearance and local details of endoscopic images for EGJA staging diagnosis. The performance of our model is assessed using accuracy, sensitivity, specificity, positive predictive value, negative predictive value, area under the receiver operating characteristic curve, average precision, and Kappa. 30 endoscopists with varying experience levels were recruited for comparative and AI-assisted evaluations. 

\textbf{Findings:} Our model demonstrates satisfactory performance for EGJA staging diagnosis across three test sets, achieving an accuracy of 0.9256 (95\% CI 0.9086-0.9426), 0.8895 (95\% CI 0.8739-0.9052), and 0.8956 (95\% CI 0.8813-0.9112), respectively. In contrast, among representative AI models, the best one (ResNet50) achieves an accuracy of 0.9125 (95\% CI 0.8942-0.9308), 0.8382 (95\% CI 0.8198-0.8566), and 0.8519 (95\% CI 0.8345-0.8693) on the three test sets, respectively; the expert endoscopists achieve an accuracy of 0.8147 (95\% CI 0.7895-0.8399) on the held-out test set. Statistical analysis reveals that our model significantly outperforms representative AI models and endoscopists (all P < 0.05), with the exception of ResNet50 on the held-out test set (P = 0.54). Moreover, with the assistance of our model, the overall accuracy for the trainee, competent, and expert endoscopists improves from 0.7035 (95\% CI 0.6739-0.7331), 0.7350 (95\% CI 0.7064-0.7636), and 0.8147 (95\% CI 0.7895-0.8399) to 0.8497 (95\% CI 0.8265-0.8728), 0.8521 (95\% CI 0.8291-0.8751), and 0.8696 (95\% CI 0.8478-0.8914), respectively. 

\textbf{Interpretation:} To our knowledge, our model is the first application of foundation models for EGJA staging diagnosis and demonstrates great potential in both diagnostic accuracy and efficiency. Besides, the results by the developed model can also be visually probed and interpreted, highlighting the clinical benefits in precision therapy. Nevertheless, the study also has limitations such as the regional constraint of the data sources and the restriction to white-light and narrow-band imaging modalities, which may limit the generalisability of the developed model. 

\textbf{Funding:} This study was supported by the Shanghai Health Development Commission, Shanghai Science and Technology Commission, Tongji University, and Shanghai Municipal Commission of Economy and Informatization.

\end{abstract}

\end{frontmatter}

\section{Introduction}\label{Sec:intro}

Esophagogastric junction adenocarcinoma (EGJA) is defined as a malignant tumour with its centre located within 5.0 centimetres proximal or distal to the esophagogastric junction, and it crosses or contacts the esophagogastric junction. It emerges as a rare yet often lethal condition that has become a significant public health issue in recent decades \cite{ref1}. The incidence of EGJA has shown a significant upward trend globally \cite{ref2,ref3}. Its 5-year survival rate is less than 5\% when patients with EGJA are diagnosed at an advanced stage, but the rate is as high as 90\% if patients are detected at an early stage \cite{ref4}. Therefore, its early diagnosis is essential for patient prognosis \cite{ref5}.

Endoscopy is the key examination for EGJA diagnosis. Although radiological imaging techniques such as computed tomography and magnetic resonance imaging have played a significant role in medical diagnosis, they often appear ineffective in identifying the early stage of EGJA \cite{ref6}. In contrast, endoscopic examination can provide intuitive visual observation, enabling endoscopists to identify early lesions more effectively. Notwithstanding, diagnosing the early stage of EGJA requires high proficiency from endoscopists. The main difficulties lie in the exposure of the esophagogastric junction, the asymptomatic manifestations of early lesions, and the inconspicuous changes in the mucosa \cite{ref7}. To improve the accuracy, various endoscopic imaging techniques have been introduced \cite{ref8}, such as chromoendoscopy, narrow-band imaging (NBI), and high-magnification endoscopy. These techniques, despite being effective, also come with certain drawbacks, such as high costs and limited accessibility. Besides, the accuracy of endoscopic screening is inextricably dependent on the experience level of endoscopists. A meta-analysis study involving 3,787 patients with esophageal or gastric cancer reports that 11.3\% of patients are missing at endoscopy within 3 years before diagnosis \cite{ref9}. This situation will only be exacerbated for EGJA due to the unique anatomy of the esophagogastric junction.

Artificial intelligence (AI) techniques have been rapidly advanced over the last decade, enabling fast and accurate disease diagnosis in automation \cite{ref10}. In the field of endoscopy, an increasing number of AI models have been developed to improve the diagnostic accuracy of gastrointestinal diseases \cite{ref11,ref12,ref13,ref14,ref15}. Most of them are based on the convolutional neural network (CNN) \cite{ref16}. Yet their performance has not reached a satisfactory level for EGJA diagnosis. Besides, AI models capable of the staging diagnosis of EGJA are still lacking \cite{ref17,ref18}. For instance, Iwagami et al.\cite{ref18} proposed a 16-layer CNN but only achieved the distinction between normal tissues and tumours without addressing the staging of EGJA.

Recently, vision foundation models such as SAM \cite{ref19} and DINO \cite{ref20} have gained significant attention in AI research, achieving state-of-the-art performance on a wide range of downstream tasks. Many studies have demonstrated their effectiveness in various medical imaging applications, such as medical report generation \cite{ref21}, medical image registration \cite{ref22}, and classification \cite{ref23}. Nevertheless, applying them directly to medical tasks tends to show certain performance degradation, due to the fact that the medical imaging modalities differ significantly from those of natural images \cite{ref24}. Besides, patient privacy and confidentiality further make it difficult to collect sufficient data to fine-tune the models, leading them to underperform CNNs.

In summary, although some promising results have been published regarding the use of AI in the diagnosis of EGJA, their clinical significance still needs to be enhanced. To address it, this study aims to develop an effective AI model for EGJA staging diagnosis, which is the first attempt of its kind. More specifically, we choose DINOv2 (a transformer-based vision foundation model \cite{ref25})  and ResNet50 (a typical CNN \cite{ref26}) as two encoders for extracting the global appearance and local details of endoscopic images, which are then fused via an element-wise gating network in a mixture-of-experts (MoE) architecture \cite{ref27} to obtain a robust feature representation for EGJA classification (early EGJA, advanced EGJA, and control).

Last but not the least, to make this study persuasive, we build the largest endoscopic EGJA dataset (12,302 images from 1,546 patients) from seven hospitals, comprising a training set, a held-out test set, an external test set, and a prospective test set. Experiment results demonstrate our model significantly outperforms both representative AI models and endoscopists with varying experience levels. Especially, the model’s strong performance on the external and prospective test sets proves its generalisability and robustness; whilst the AI-assisted evaluation and visual analysis showcase the clinical applicability of our model and enhance its interpretability.

\section{Methods}

\subsection{Ethics}

This study was approved by the ethics board of Tongji Hospital Affiliated to Tongji University (Approval No.2023-025) and was endorsed by the participating hospitals. For the retrospective study part, the requirement for informed consent from participants was waived by ethics committees. For the prospective study part, all patients signed the informed consent. Additionally, this study was reported in accordance with the TRIPOD+AI guidelines \cite{ref28}.

\subsection{Ground truth definition for EGJA}

It is challenging to obtain the true EGJA staging diagnosis solely based on the endoscopic tumour characteristics. Therefore, we employ the pathological evaluation of intact EGJA lesions as the golden standard to classify EGJA into two categories: early EGJA (E-EGJA) and advanced EGJA (A-EGJA). According to the World Health Organization Classification for EGJA, the pathological diagnostic criteria for E-EGJA refer to high-grade dysplasia (Tis) and tumour invasion into the lamina propria, muscularis mucosae, or submucosa (T1), with no evidence of lymphovascular invasion. The pathological diagnostic criteria for A-EGJA refer to tumour invasion beyond the submucosa ($\geq$T2) accompanied by lymph node or distant metastasis.

\subsection{Patients and image quality control}

\begin{figure*}[!t]
\begin{center}
\begin{tabular}{c}
\includegraphics[width=6in]{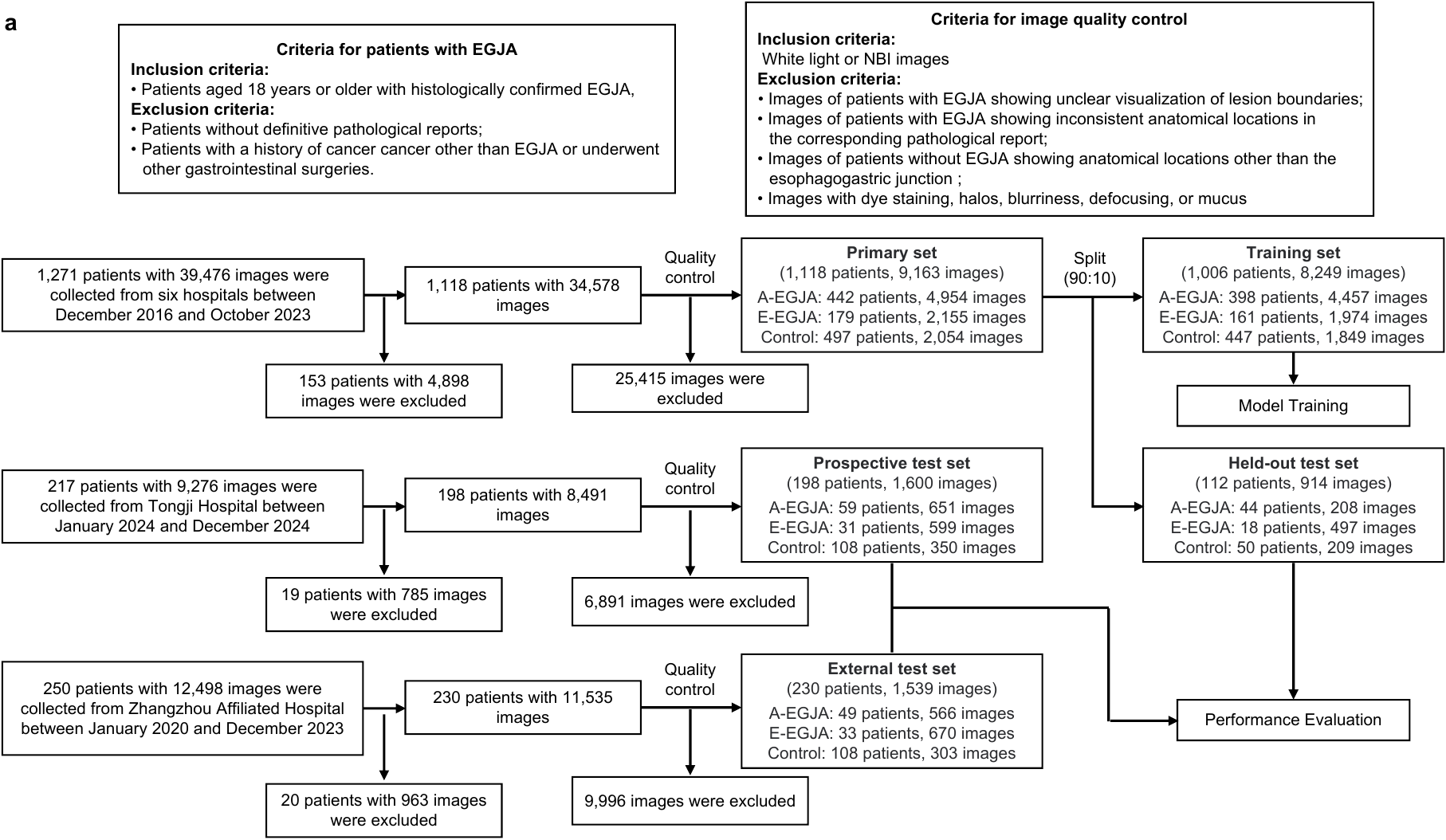} \\
\includegraphics[width=6in]{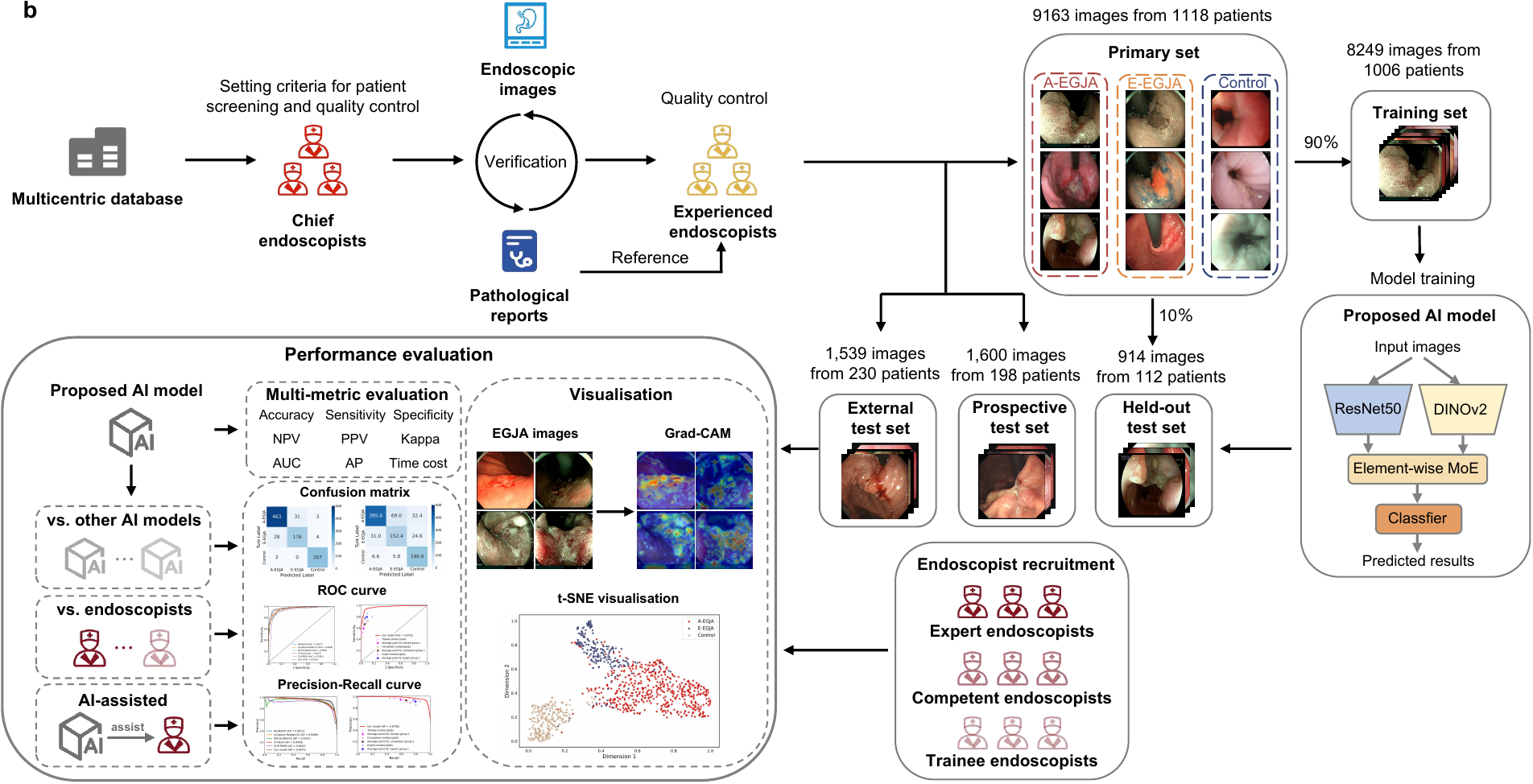}  \\
\end{tabular}
\end{center}
\caption{Flowchart of patients and study design. \textbf{a:} the patient screening and image quality control process for each image set used for model training and performance evaluation. \textbf{b:} the development and evaluation of our proposed AI model to achieve EGJA staging diagnosis.}
\label{fig1}
\end{figure*}

1,738 patients (with or without EGJA) with 61,250 images were initially collected from the seven hospitals (Renji Hospital, Huashan Hospital, Zhongshan Hospital, Xinhua Hospital, Ruijin Hospital, Tongji Hospital, and Zhangzhou Affiliated Hospital of Fujian Medical University) in a time span between December 28, 2016 and December 30, 2024. Each patient underwent upper gastrointestinal endoscopy, at least 25 images and up to 50 images were captured originally. Three chief endoscopists with more than 15 years of endoscopic experience developed the protocol for patient screening and image quality control (Fig.~\ref{fig1}(a)).

During the screening stage, patients aged 18 years or older with histologically confirmed EGJA are eligible for this study. They are excluded if meeting any of the following criteria: (1) with a history of cancer other than EGJA or underwent other gastrointestinal surgeries; (2) without definitive pathological reports. All pathological assessments are based on hematoxylin-stained and eosin-stained tissue slides. For patients without EGJA (e.g., healthy cardia mucosa or benign conditions), there are no specific exclusion criteria.

After patient screening, each patient with multiple images is assigned a single and unified label (E-EGJA, A-EGJA, or control) based on the pathological report. Eight experienced endoscopists (each with $\geq$5 years of experience) assess the image quality of all patients according to the protocol developed by the three expert endoscopists. Specifically, white light or NBI images are generally considered for inclusion. They are excluded if they meet any of the following criteria: (1) images of patients with EGJA showing unclear lesion boundaries; (2) images of patients with EGJA showing inconsistent anatomical locations in the corresponding pathological report; (3) images of patients without EGJA showing anatomical locations other than the esophagogastric junction; (4) images of all patients with dye staining, halos, blurriness, defocusing, or mucus. To improve the credibility of the data, the eight endoscopists are divided into four groups, each comprising two endoscopists. The endoscopists in each group need to reach a consensus on images assigned to them. Any images with divergent views are also excluded. As a result, 12,302 well-labelled images with high quality from 1546 patients can be utilised for our dataset construction.

\subsection{Dataset construction}

As shown in Fig.~\ref{fig1}, the primary set comprises 9,163 images from 1,118 patients collected from six hospitals between December 2016, and October 2023. It comprises 2,155 E-EGJA images from 179 patients, 4,954 A-EGJA images from 442 patients, and 2,054 control images from 497 patients. Then, the primary set is divided into the training set (4,457 A-EGJA images from 398 patients, 1,947 E-EGJA images from 161 patients, 1845 control images from 447 control patients), and the held-out test set (497 A-EGJA images from 44 patients, 208 E-EGJA images from 18 patients, 209 control images from 50 patients). Notably, there is no overlap among patients on these two sets, ensuring the patient-level data integrity. Apart from the held-out test set, we build an external test set and a prospective test set to comprehensively evaluate the generalisability of our model. The external test set comprises 1,539 images (566 A-EGJA, 670 E-EGJA, 303 control) from 230 patients collected from Zhangzhou Affiliated Hospital of Fujian Medical University between January 2020, and December 2023. The prospective test set comprises 1,600 images (651 A-EGJA, 599 E-EGJA, 350 control) from 198 patients who made their first visit to the Tongji Hospital between January 2024 and December 2024. Detailed patient demographics are presented in Table~\ref{table1}.

\begin{table*}[!t]
\caption{{Distribution of patient demographics in our collected image sets.}}
%\vspace{-0.1in}
\label{table1}
\begin{center}
\footnotesize
\setlength{\tabcolsep}{1.5mm}
\begin{tabular}{ccccc}
\toprule
& \textbf{Training set} & \textbf{Held-out test set} & \textbf{External test set} & \textbf{Prospective test set} \\ \midrule
Total patients & 1,006 & 112   & 230   & 198 \\ \addlinespace[0.4em]
Total images & 8,249 & 914   & 1539  & 1600 \\ \addlinespace[0.4em]
Age (years, mean ± SD) & 63.5 ± 11.2 & 64.1 ± 10.8 & 69.5 ± 8.9 & 68.9 ± 9.4 \\ \addlinespace[0.4em]
Male  & 676   & 74    & 181   & 153 \\ \addlinespace[0.4em]
Female & 330   & 38    & 49    & 45 \\ \addlinespace[0.4em]
Patients with A-EGJA & 398   & 44    & 89    & 59 \\ \addlinespace[0.4em]
A-EGJA images & 4,457 & 497   & 566   & 651 \\ \addlinespace[0.4em]
Patients with E-EGJA & 161   & 18    & 33    & 31 \\ \addlinespace[0.4em]
E-EGJA images & 1,947 & 208   & 670   & 599 \\ \addlinespace[0.4em]
Patients without EGJA & 447   & 50    & 108   & 108 \\ \addlinespace[0.4em]
Control images & 1,845 & 209   & 303   & 350 \\
\bottomrule
\end{tabular}
\end{center}
\end{table*}

\subsection{Multicentre data standardisation}

Our training data were collected from 6 hospitals. To address the centre-specific variability, we have developed systematic patient screening and image quality control protocol to make sure only those EGJA images with clear lesions and consistent anatomical locations from each hospital are included into the dataset. Moreover, several essential data standardisation steps, including image normalisation and image augmentation, are conducted. For the former, we calculate the global statistics (mean  and standard deviation  over all images) to normalise all the images before the model training: , where  represents the original image and  represents the normalised image. For the latter, we simultaneously perform random cropping, rotation, and colour jittering on each image before the model training. These are common tracks used in multicentre studies as they serve the function to mitigate centre-specific variability and improve model’s generalisability \cite{ref29,ref30,ref31}. An ablation study for these standardisation steps is provided for both our default model (Ours, See the Model description Section) and its two variants (Ours-v1 and Ours-v2, See the Comparison between image-level training and patient-level training Section) in the Supplementary Material.

\subsection{Model description}

Since CNN is well-known for capturing local features while the transformer-based foundation model excels at extracting global features, we attempt to reinforce the complementary advantages of both architectures for EGJA staging diagnosis. As depicted in Supplementary Fig.~\ref{sup:fig1}, our model mainly consists of two encoder branches (DINOv2 \cite{ref25} and ResNet50 \cite{ref26}), a gating network to implement MoE \cite{ref27}, and a classifier. The DINOv2 and ResNet50 encoders are employed to extract features of global appearance and local details of the input image, respectively. Then, a gating network is built to implement MoE for the element-wise feature fusion. Finally, the fused feature is fed into the classifier that outputs the predicted result for the input image. The detailed description of our model is provided in the Supplementary Material.

\subsection{The training configurations of our model}

Following the common practice in state-of-the-art endoscopic disease classification works \cite{ref31,ref32,ref33,ref34}, our default model is trained at the image level by treating multiple images from the same patient as independent observations. We implement our model under the PyTorch framework and deploy it using two Tesla A40 GPUs. During model training, a ResNet50 with its pre-trained parameters on the ImageNet dataset \cite{ref35} is fine-tuned on our training set for 40 epochs in advance, after which the entire model including the DINOv2 encoder (pre-trained on large-scale natural images) is fine-tuned for another 40 epochs. Cross entropy loss is applied as the loss function. The Adam optimiser is employed with an initial learning rate of 0.0001. All endoscopic images are uniformly resized to 448×448 as model inputs. The batch size is set to 128.

\subsection{Comparison between image-level training and patient-level training}

To validate the necessity of the default image-level training paradigm, we present two variants of our model, Ours-v1 and Ours-v2, by fusing image features from multiple images of the same patient into one feature representation for patient-level training. Specifically, both models extract features from images of the same patient using the same feature extractor as our default model. Then, Ours-v1 performs the max-pooling operation to fuse these features into one feature representation while Ours-v2 fuses these features through a 3-layer Bi-directional Long Short-Term Memory (BiLSTM). The fused feature is fed into the classifier to predict a result for this patient. We train both models on the same training data as our default model and test them on the held-out test set at the patient level. During inference, multiple images of the same patient are fed into Ours-v1 or Ours-v2 to directly obtain one patient-level diagnostic output, while our default model obtains this output by averaging the predicted probabilities across images of the same patient.

\subsection{Comparison with other AI models}

To demonstrate the effectiveness of our model, we compare it against six representative AI models, including three CNNs (ResNet50 \cite{ref26}, Inception-ResNet-V2 \cite{ref36}, EfficientNetV2 \cite{ref37}) and three vision foundation models (ViT-B/16 \cite{ref38}, CLIP-RN50 \cite{ref39}, DINOv2-S/14 \cite{ref25}). All models adopt the default image-level training paradigm. The five-fold cross validation is conducted for each model. Specifically, the training set is randomly divided into five mutually exclusive subsets. Each subset is successively employed as the validation set while the remaining four subsets are employed for model training. Then, the best performing model is selected to evaluate on the test sets.

\subsection{Endoscopists’ performance with and without our model’s assistance}

30 endoscopists with varying levels of experience are recruited to evaluate the clinical applicability of our model on the held-out test set. Note that all these endoscopists are not involved in the process of patient screening and image quality control. For the sake of convenience, we classify them into three categories, namely trainee (with less than 3 years of experience), competent (with 3 to 10 years of experience), or expert (with more than 10 years of experience) endoscopists. Each category has 10 endoscopists. Then, all endoscopists are divided into 6 groups, namely trainee group A/B, competent group A/B, and expert group A/B, each with 5 endoscopists corresponding to their levels of expertise. Endoscopists in each group A independently make their diagnosis for each image on the test sets while endoscopists in each group B make their diagnosis with the assistance of our model’s prediction.

\subsection{Model interpretability}

We employ the gradient-weighted class activation maps (Grad-CAM \cite{ref40}) and t-distributed stochastic neighbor embedding (t-SNE \cite{ref41}) to investigate the interpretability of our model. The Grad-CAM can visualise the regions of interest for our model in diagnosing EGJA, which can be used to analyse the key diagnostic features of E-EGJA and A-EGJA under endoscopy. We select some representative cases from the held-out, external, and prospective test sets to perform the Grad-CAM for visual analysis. The t-SNE can visualise the image features by embedding them into points in the feature space, which can reveal the cluster separation between E-EGJA, A-EGJA, and control images. We apply t-SNE to the fused feature  of our model to obtain the two-dimensional embeddings of each image across three test sets and visualise the embeddings separately for each set.

\subsection{Statistical analysis}

To assess the effectiveness of our model, the accuracy, sensitivity, specificity, positive predictive value (PPV), negative predictive value (NPV) are calculated with the 95\% confidence interval (CI) \cite{ref42}. The receiver operating characteristic (ROC) curve is created via plotting sensitivity against specificity while the precision-recall curve is created via plotting precision (equivalent to PPV) against recall (equivalent to sensitivity). The area under the curve (AUC) and average precision (AP) are obtained by calculating the area under the ROC and precision-recall curves, respectively. For overall comparison in all test sets, the sensitivity, specificity, PPV, and NPV are computed as the average across all categories. AUC and AP are calculated as the micro average. Kappa is calculated based on the confusion matrix, which measures the agreement between the prediction and the ground truth. The time cost is provided as the average processing time for a single image in the test set. 

Statistical analysis is conducted using the McNemar-Bowker test, DeLong test and Kappa consistency test. The McNemar-Bowker test is conducted to assess the classification differences between our model and different AI models or endoscopist groups (Table~\ref{table2}-\ref{table6} and Supplementary Table~\ref{sup:tab3}-\ref{sup:tab5}); the DeLong test is conducted to compare the AUC of our model with different AI models (Fig.~\ref{fig2}d and Supplementary Table~\ref{sup:tab6}); the Kappa consistency test is conducted to compare the diagnostic consistency between our model and different endoscopist groups (Fig.~\ref{fig3}a). P < 0.05 is regarded as statistically significant. All statistical analyses are performed using Python (version 3.9).

To comprehensively evaluate our model’s performance, we conduct both image-level and patient-level evaluations utilising the above metrics and analyses. For the former, each image is independently classified by our model and is evaluated against its corresponding ground truth. Experiments are conducted on the held-out, external, and prospective test sets. For the latter, we average the predicted probabilities over images of the same patient to obtain the patient-level diagnostic output by the model, which is evaluated against the corresponding ground truth per patient. Experiments are conducted on the held-out test set. Notably, since the number of images varies per patient, this clustering effect may cause model overfitting. While the current image-level evaluation does not account for the clustering effect per patient, hence it may not sufficiently reflect the possible model overfitting. To cope with it, we also conduct a weighted image-level evaluation: each image’s result is weighted by the inverse of the number of images from the same patient.

\subsection{Role of the funding source}

The funding source had no role in the study design, data collection, data analysis, data interpretation, or writing of the report. The corresponding authors had access to all the data in the study and had final responsibility for the decision to submit the manuscript for publication.

\section{Results}
\subsection{Model performance at the image level}

The flow chart of this study is illustrated in Fig.~\ref{fig1}. Supplementary Table~\ref{sup:tab1} and \ref{sup:tab2} present the average results of each AI model after training on the training set (n=8249) using five-fold cross validation. Subsequently, the best performing models are evaluated on the held-out (n=914), external (n=1539), and prospective (n=1600) test sets. Table~\ref{table2} summarises the overall performance of all AI models; our model achieves the best performance amongst others, with the highest accuracy (0.9256, 0.8895, 0.8963) on the three test sets. The P values (calculated by the McNemar-Bowker test) indicate that the performance of our model is statistically different from other AI models (P<0.05), except for ResNet50 on the held-out test set (P=0.54). Detailed results across categories (A-EGJA, E-EGJA, and control) are provided in Supplementary Table~\ref{sup:tab3}-\ref{sup:tab5}.

\begin{figure}[!t]
\begin{center}
\begin{tabular}{c}
\includegraphics[width=5.7in]{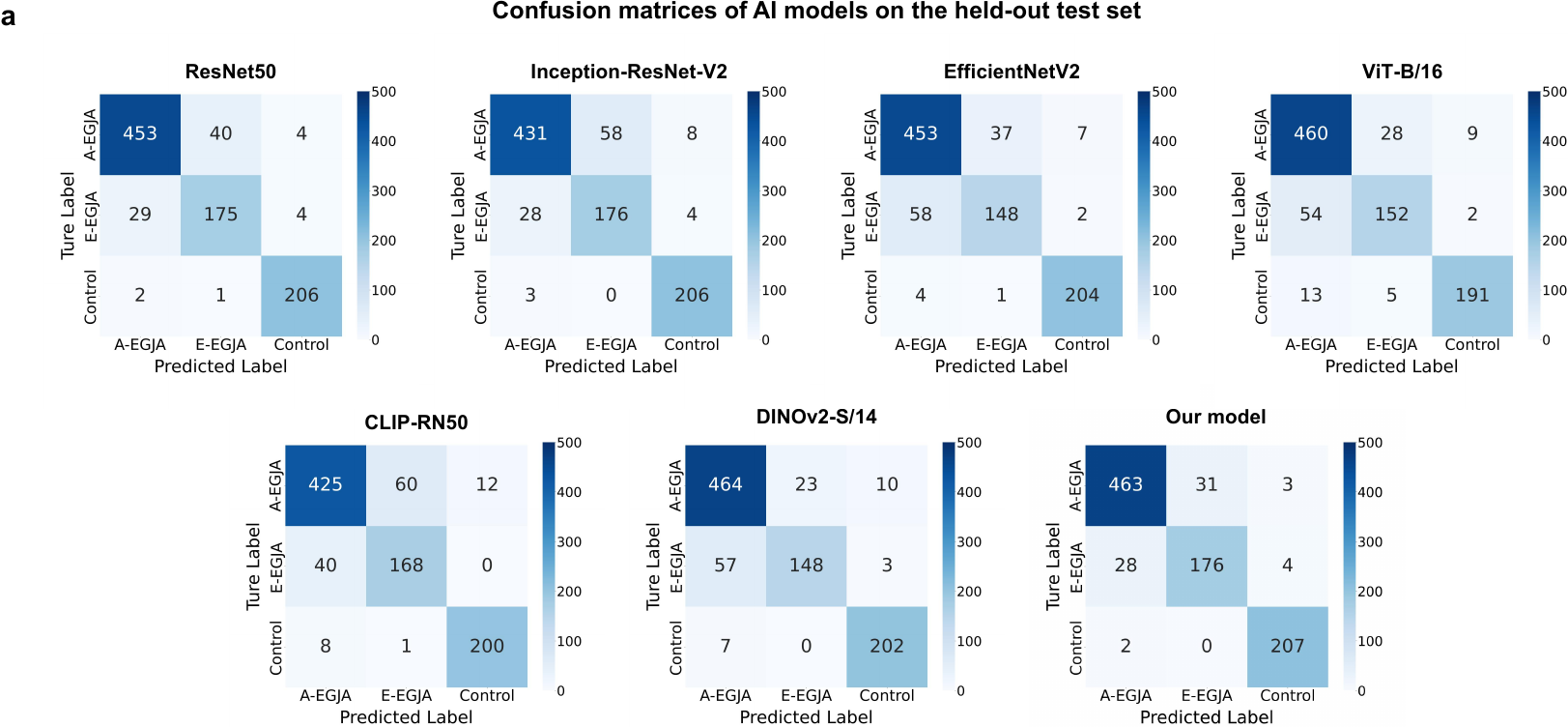} \\
\includegraphics[width=5.7in]{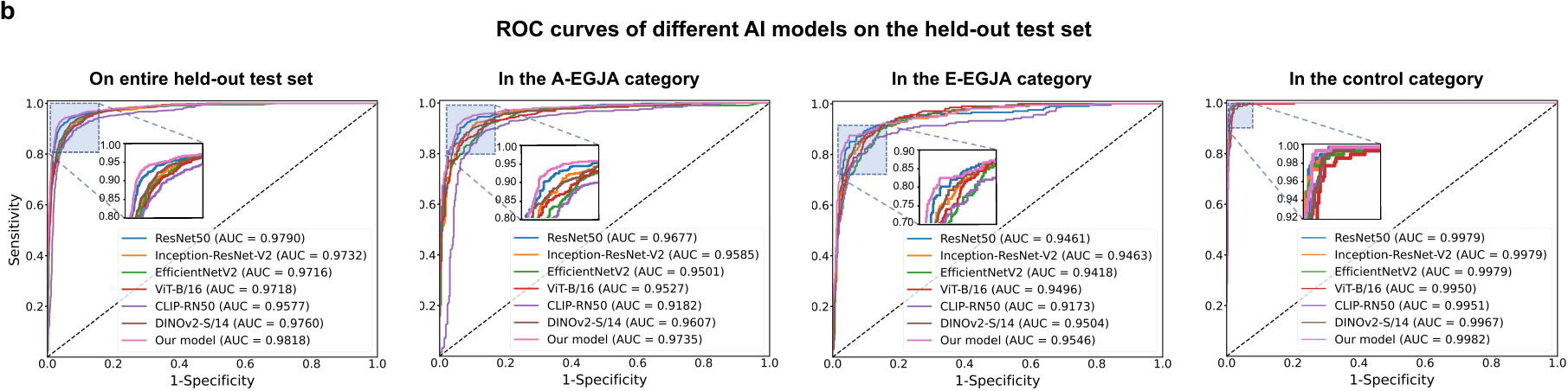}  \\
\includegraphics[width=5.7in]{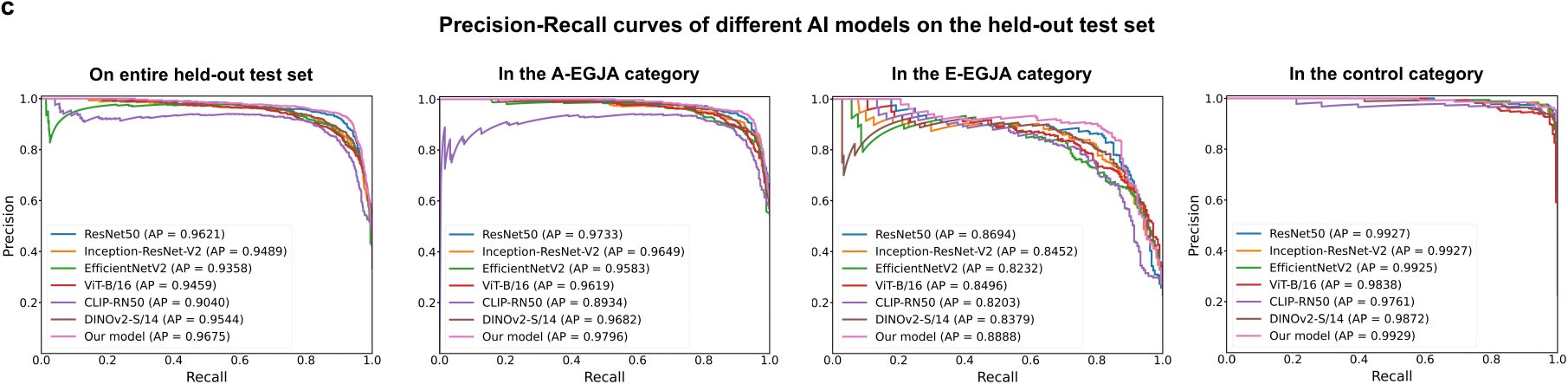} \\
\includegraphics[width=5.7in]{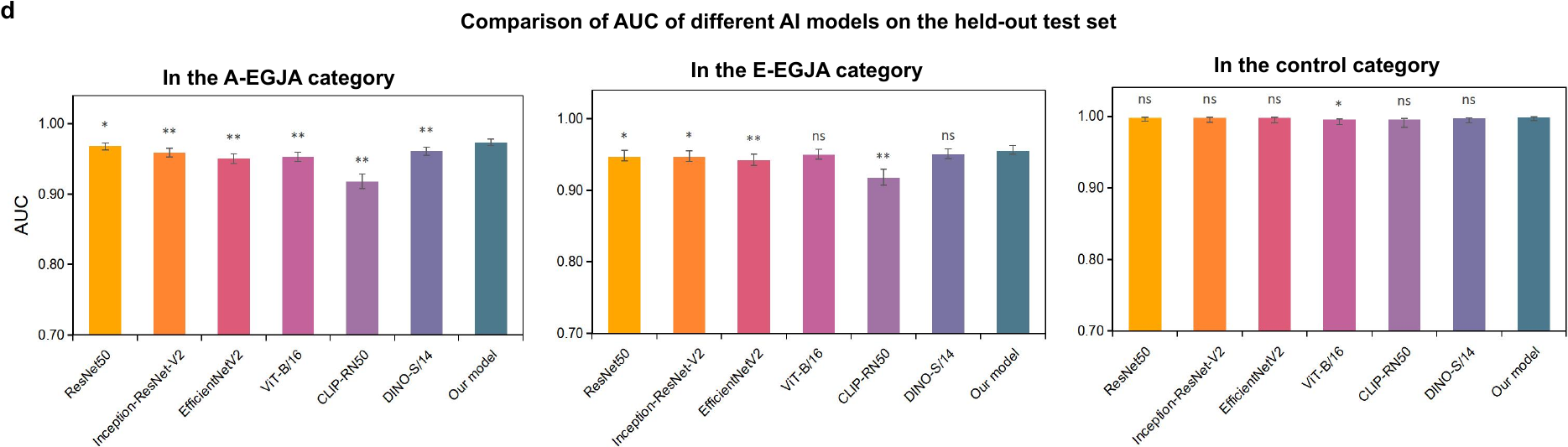}  \\
\end{tabular}
\end{center}
\setlength{\abovecaptionskip}{-5pt}
\caption{Classification performance of different AI models on the held-out test set. \textbf{a:} the classification confusion matrices of different models. Numbers in each matrix represent the count of images in each category classified correctly (diagonal) and incorrectly (off the diagonal). \textbf{b:} the ROC curves of different models. The first column shows the ROC curves on the entire test set while the second to last columns represent the curves across categories (A-EGJA, E-EGJA, control) of the test set. \textbf{c:} the precision-recall curves of different models. The first column shows the precision-recall curves on the entire test set while the second to last columns represent the curves across categories of the test set, respectively. \textbf{d:} the comparison of different AI models' AUC across categories. * indicates P < 0.05, ** indicates P < 0.01, and ns indicates P > 0.05 for comparison of AUC with our model using the Delong test.}
\label{fig2}
\end{figure}

\begin{table*}[!t]
\caption{{Overall comparison to representative AI models on the held-out, external, and prospective test sets at the image level.}}
%\vspace{-0.1in}
\label{table2}
\begin{center}
\scriptsize
\setlength{\tabcolsep}{1.5mm}
    \begin{tabular}{cccccccc}
    \toprule
          & \textbf{ResNet50} & \textbf{Inception-ResNet-V2} & \textbf{EfficientNetV2} & \textbf{ViT-B/16} & \textbf{CLIP-RN50} & {\textbf{DINOv2-S/14}} & \textbf{Our model} \\
    \midrule
    \multicolumn{8}{c}{\textbf{Held-out test set (n=914)}} \\ \addlinespace[0.4em]
    Accuracy & 0.9125 & 0.8895 & 0.8807 & 0.8786 & 0.8676 & 0.8906 & \textbf{0.9256} \\
    (95\% CI) & (0.8942-0.9308) & (0.8692-0.9098) & (0.8597-0.9018) & (0.8574-0.8997) & (0.8456-0.8896) & (0.8704-0.9108) & (0.9086-0.9426) \\ \addlinespace[0.4em]
    Sensitivity & 0.9128 & 0.8997 & 0.8664 & 0.8567 & 0.8733 & 0.8705 & \textbf{0.9227} \\
    (95\% CI) & {(0.8826-0.9431)} & {(0.8680-0.9313)} & (0.8306-0.9021) & (0.8163-0.8972) & (0.8359-0.9106) & {(0.8346-0.9065)} & (0.8946-0.9509) \\ \addlinespace[0.4em]
    {Specificity} & 0.9521 & 0.9422 & 0.9282 & 0.9257 & 0.9272 & 0.9318 & \textbf{0.9581} \\
    {(95\% CI)} & (0.9353-0.9688) & (0.9238-0.9605) & (0.9085-0.9479) & (0.9057-0.9456) & (0.9069-0.9475) & {(0.9126-0.951)} & (0.9423-0.9738) \\ \addlinespace[0.4em]
    {PPV} & 0.9029 & 0.8767 & 0.8777 & 0.88  & 0.8585 & 0.8946 & \textbf{0.9189} \\
    {(95\% CI)} & (0.8697-0.9361) & (0.8405-0.9128) & (0.8400-0.9154) & (0.8417-0.9183) & (0.8200-0.8970) & {(0.8577-0.9316)} & (0.8877-0.9501) \\ \addlinespace[0.4em]
    {NPV} & 0.9487 & 0.9342 & 0.9334 & 0.9341 & 0.9218 & 0.9412 & \textbf{0.957} \\
    {(95\% CI)} & (0.9323-0.9651) & (0.9164-0.9520) & (0.9144-0.9524) & (0.9140-0.9542) & (0.9017-0.9420) & {(0.923-0.9595)} & (0.9419-0.9721) \\ \addlinespace[0.4em]
    Kappa & 0.8553 & 0.8195 & 0.7992 & 0.7941 & 0.7825 & 0.8144 & \textbf{0.8763} \\ \addlinespace[0.4em]
    AUC   & 0.979 & 0.9732 & 0.9716 & 0.9718 & 0.9577 & 0.976 & \textbf{0.9818} \\ \addlinespace[0.4em]
    AP    & 0.9621 & 0.9489 & 0.9357 & 0.9459 & 0.904 & 0.9544 & \textbf{0.9675} \\ \addlinespace[0.4em]
    P value & 0.54  & 0.0017 & 0.024 & 0.0055 & 0.017 & {<0.0001} & Reference \\ \midrule
    \multicolumn{8}{c}{\textbf{External test set (n=1539)}} \\ \addlinespace[0.4em]
    Accuracy & 0.8382 & 0.8278 & 0.8135 & 0.8252 & 0.7947 & 0.8324 & \textbf{0.8895} \\
    (95\% CI) & (0.8198-0.8566) & (0.8089-0.8467) & (0.7941-0.8330) & (0.8062-0.8442) & (0.7745-0.8149) & {(0.8137-0.8510)} & (0.8739-0.9052) \\ \addlinespace[0.4em]
    Sensitivity & 0.8432 & 0.8383 & 0.8209 & 0.8291 & 0.8001 & 0.8334 & \textbf{0.9004} \\
    (95\% CI) & (0.8110-0.8754) & (0.8063-0.8702) & (0.7868-0.8550) & (0.7954-0.8628) & (0.7643-0.8359) & {(0.7999-0.8670)} & (0.8748-0.9259) \\ \addlinespace[0.4em]
    Specificity & 0.9124 & 0.9096 & 0.9009 & 0.9045 & 0.8912 & 0.9115 & \textbf{0.9408} \\
    (95\% CI) & (0.8955-0.9294) & (0.8920-0.9272) & (0.8828-0.9190) & (0.8873-0.9218) & (0.8727-0.9097) & {(0.8943-0.9287)} & (0.9267-0.9549) \\ \addlinespace[0.4em]
    PPV   & 0.8454 & 0.8239 & 0.8155 & 0.8399 & 0.8042 & 0.8334 & \textbf{0.8953} \\
    (95\% CI) & (0.8133-0.8776) & (0.7899-0.8579) & (0.7808-0.8501) & (0.8080-0.8718) & (0.7692-0.8392) & {(0.8000-0.8667)} & (0.8686-0.9220) \\ \addlinespace[0.4em]
    NPV   & 0.9133 & 0.9084 & 0.9001 & 0.9057 & 0.8906 & 0.9109 & \textbf{0.9397} \\
    (95\% CI) & (0.8964-0.9303) & (0.8914-0.9255) & (0.8821-0.9180) & (0.8880-0.9234) & (0.8719-0.9094) & {(0.8936-0.9281)} & (0.9258-0.9537) \\ \addlinespace[0.4em]
    Kappa & 0.7453 & 0.7311 & 0.7078 & 0.7242 & 0.6783 & 0.737 & \textbf{0.827} \\ \addlinespace[0.4em]
    AUC   & 0.9565 & 0.9419 & 0.9456 & 0.9357 & 0.9072 & 0.9391 & \textbf{0.9701} \\ \addlinespace[0.4em]
    AP    & 0.92  & 0.8965 & 0.9066 & 0.8835 & 0.8438 & 0.8959 & \textbf{0.9471} \\ \addlinespace[0.4em]
    P value & 0.0019 & <0.0001 & 0.0061 & 0.0016 & 0.015 & 0.038 & Reference \\ \midrule
    \multicolumn{8}{c}{\textbf{Prospective test set (n=1600)}} \\ \addlinespace[0.4em]
    Accuracy & 0.8519 & 0.8331 & 0.83  & 0.8344 & 0.8056 & 0.8356 & \textbf{0.8963} \\
    (95\% CI) & (0.8345-0.8693) & (0.8149-0.8514) & (0.8116-0.8484) & (0.8162-0.8526) & (0.7862-0.825) & {(0.8175-0.8538)} & (0.8813-0.9112) \\ \addlinespace[0.4em]
    Sensitivity & 0.8655 & 0.8473 & 0.8405 & 0.8426 & 0.8089 & 0.8437 & \textbf{0.9052} \\
    (95\% CI) & (0.8376-0.8934) & (0.8177-0.877) & (0.8095-0.8714) & (0.8115-0.8736) & (0.7748-0.8429) & {(0.8128-0.8745)} & (0.8813-0.9292) \\ \addlinespace[0.4em]
    Specificity & 0.9227 & 0.9148 & 0.912 & 0.9129 & 0.8987 & 0.9135 & \textbf{0.9462} \\
    (95\% CI) & (0.9069-0.9386) & (0.898-0.9316) & (0.895-0.929) & (0.8963-0.9294) & (0.881-0.9164) & {(0.8968-0.9301)} & (0.9327-0.9598) \\ \addlinespace[0.4em]
    PPV   & 0.8545 & 0.8303 & 0.8297 & 0.8415 & 0.808 & 0.8401 & \textbf{0.8956} \\
    (95\% CI) & (0.8242-0.8847) & (0.798-0.8626) & (0.7972-0.8622) & (0.8103-0.8727) & (0.7742-0.8418) & {(0.8085-0.8717)} & (0.8691-0.9221) \\ \addlinespace[0.4em]
    NPV   & 0.9214 & 0.9126 & 0.9106 & 0.9126 & 0.8985 & 0.9134 & \textbf{0.945} \\
    (95\% CI) & (0.906-0.9367) & (0.8965-0.9287) & (0.894-0.9272) & (0.8961-0.9292) & (0.8806-0.9165) & {(0.8969-0.9298)} & (0.9319-0.958) \\ \addlinespace[0.4em]
    Kappa & 0.7721 & 0.7443 & 0.7385 & 0.7442 & 0.6998 & 0.7461 & \textbf{0.8402} \\ \addlinespace[0.4em]
    AUC   & 0.9645 & 0.9477 & 0.9491 & 0.9408 & 0.8999 & 0.9486 & \textbf{0.9743} \\ \addlinespace[0.4em]
    AP    & 0.9375 & 0.9032 & 0.9083 & 0.887 & 0.8265 & 0.9057 & \textbf{0.9551} \\ \addlinespace[0.4em]
    P value & 0.034 & <0.0001 & 0.0019 & 0.0006 & <0.0001 & 0.038 & Reference \\
    \bottomrule
    \end{tabular}%
\end{center}
\end{table*}

Next, the confusion matrices (Fig.~\ref{fig2}a, Supplementary Fig.~\ref{sup:fig2}a and \ref{sup:fig3}a) demonstrate that our model exhibits superior robustness on the three test sets, with the biggest numbers on the matrix diagonals. Comparisons of the ROC and precision-recall curves among the AI models (Fig.~\ref{fig2}b and c, Supplementary Fig.~\ref{sup:fig2}b and c, and Supplementary Fig.~\ref{sup:fig3}b and c) reveal that our model dominates in both ROC and precision-recall spaces, thereby achieving the highest AUC (0.9818, 0.9701, 0.9743) and AP (0.9675, 9471, 0.9551) on the three test sets. Furthermore, Fig.~\ref{fig2}d, Supplementary Fig.~\ref{sup:fig2}d and \ref{sup:fig3}d visualise the AUC comparison of different AI models across all categories on the three test sets, and Supplementary Table~\ref{sup:tab6} provides their P values (calculated by the DeLong test) compared with our model. We observe that our model obtains the highest AUC in each category. Particularly, in the A-EGJA and E-EGJA categories, the AUC of our model is statistically higher than the majority of AI models (P<0.05) in the E-EGJA category on the held-out test set. Consequently, all these results consistently support our model’s strong capability and generalisability for EGJA screening and staging diagnosis.

Additionally, as indicated by the end of the Statistical analysis Section, in Supplementary Table~\ref{sup:tab7}, we also compare the weighted image-level results with our original results on the held-out test set: the results show that the weighted results are indeed close to our original results, indicating the robustness of our model with no significant overfitting.

\subsection{Model performance at the patient level}

\begin{table*}[!t]
\caption{{Comparison of our model trained at the image level (Ours) or patient-level (Ours-v1 and Ours-v2). Evaluation is on the held-out test set at the patient level.}}
%\vspace{-0.1in}
\label{table3}
\begin{center}
\scriptsize
\setlength{\tabcolsep}{1.5mm}
    \begin{tabular}{ccccccccc}
    \toprule
    \multirow{2}{*}{} & \textbf{Accuracy} & \textbf{Sensitivity} & \textbf{Specificity} & \textbf{PPV} & \textbf{NPV} & \multicolumn{1}{c}{\multirow{2}{*}{\textbf{Kappa}}} & \multirow{2}{*}{\textbf{AUC}} & \multicolumn{1}{c}{\multirow{2}{*}{\textbf{AP}}} \\
          & (95\% CI) & (95\% CI) & (95\% CI) & (95\% CI) & (95\% CI) &       &       &  \\
    \midrule
    \multirow{2}{*}{\textbf{Ours}} & \textbf{0.9464} & \textbf{0.9363} & \textbf{0.9719} & \textbf{0.9464} & \textbf{0.9728} & \multirow{2}{*}{\textbf{0.9136}} & \multirow{2}{*}{\textbf{0.9948}} & \multirow{2}{*}{\textbf{0.9905}} \\
          & (0.9047-0.9881) & (0.8628-1.000) & (0.9444-0.9995) & (0.8808-1.000) & (0.9440-1.000) &       &       &  \\ \addlinespace[0.4em]
    \multirow{2}{*}{\textbf{Ours-v1}} & 0.9107 & 0.9024 & 0.959 & 0.8797 & 0.9574 & \multirow{2}{*}{0.8582} & \multirow{2}{*}{0.984} & \multirow{2}{*}{0.9698} \\
          & (0.8579-0.9635) & (0.8160-0.9774) & (0.9236-0.9929) & (0.7963-0.9612) & (0.9242-0.9879) &       &       &  \\ \addlinespace[0.4em]
    \multirow{2}{*}{\textbf{Ours-v2}} & 0.8929 & 0.8653 & 0.9401 & 0.8948 & 0.9501 & \multirow{2}{*}{0.8251} & \multirow{2}{*}{0.9678} & \multirow{2}{*}{0.9441} \\
          & (0.8356-0.9501) & (0.7633-0.9673) & (0.8892-0.9848) & (0.7875-0.9836) & (0.9132-0.9870) &       &       &  \\
    \bottomrule
    \end{tabular}%
\end{center}
\end{table*}

The held-out test set is employed to evaluate model performance at the patient level. It comprises 112 patients with 914 images, and each patient has an average of 8.2 images. In Table~\ref{table3}, we compare the performance of our model trained at the image level (Ours) or patient-level (Ours-v1 and Ours-v2): Ours performs clearly better than Ours-v1 and Ours-v2. We suggest that the reason is the information loss during multi-image feature fusion in Ours-v1 and Ours-v2. In contrast, image-level training by default helps the model generalise over a wide range of images for each patient, which in turn benefits the patient-level diagnosis.

\begin{table*}[b]
\caption{{Overall comparison to representative AI models on the held-out at the patient level.}}
%\vspace{-0.1in}
\label{table4}
\begin{center}
\scriptsize
\setlength{\tabcolsep}{1.5mm}
    \begin{tabular}{cccccccc}
    \toprule
          & \textbf{ResNet50} & \textbf{Inception-ResNet-V2} & \textbf{EfficientNetV2} & \textbf{ViT-B/16} & \textbf{CLIP-RN50} & \multicolumn{1}{c}{\textbf{DINOv2-S/14}} & \textbf{Our model} \\
    \midrule
    \multicolumn{1}{c}{Accuracy} & 0.9196 & 0.9018 & 0.9107 & 0.8571 & 0.8661 & 0.9018 & \textbf{0.9464} \\
    \multicolumn{1}{c}{(95\% CI)} & (0.8693-0.9700) & (0.8467-0.9569) & (0.8579-0.9635) & (0.7923-0.922) & (0.803-0.9291) & \multicolumn{1}{c}{(0.8467-0.9569)} & (0.9047-0.9881) \\ \addlinespace[0.4em]
    \multicolumn{1}{c}{Sensitivity} & 0.9145 & 0.8884 & 0.8969 & 0.8441 & 0.8344 & 0.8902 & \textbf{0.9363} \\
    \multicolumn{1}{c}{(95\% CI)} & \multicolumn{1}{c}{(0.8178-1.000)} & \multicolumn{1}{c}{(0.7784-0.9983)} & (0.7948-0.9990) & (0.7226-0.9656) & (0.7074-0.9613) & \multicolumn{1}{c}{(0.7861-0.9943)} & (0.8628-1.000) \\ \addlinespace[0.4em]
    \multicolumn{1}{c}{Specificity} & 0.9627 & 0.9529 & 0.9578 & 0.9311 & 0.9313 & 0.9515 & \textbf{0.9719} \\
    \multicolumn{1}{c}{(95\% CI)} & (0.9289-0.9964) & (0.9153-0.9905) & (0.9220-0.9936) & (0.8865-0.9756) & (0.8752-0.9874) & \multicolumn{1}{c}{(0.9138-0.9892)} & (0.9444-0.9995) \\ \addlinespace[0.4em]
    \multicolumn{1}{c}{PPV} & 0.8916 & 0.8741 & 0.8819 & 0.8301 & 0.8344 & 0.8832 & \textbf{0.9464} \\
    \multicolumn{1}{c}{(95\% CI)} & (0.8038-0.9795) & (0.7791-0.9692) & (0.7899-0.9740) & (0.7280-0.9321) & (0.7118-0.9570) & \multicolumn{1}{c}{(0.7900-0.9763)} & (0.8808-1.000) \\ \addlinespace[0.4em]
    \multicolumn{1}{c}{NPV} & 0.9577 & 0.9489 & 0.9545 & 0.9281 & 0.9329 & 0.9501 & \textbf{0.9728} \\
    \multicolumn{1}{c}{(95\% CI)} & (0.9129-1.000) & (0.8988-0.9989) & (0.9098-0.9992) & (0.8726-0.9837) & (0.8761-0.9897) & \multicolumn{1}{c}{(0.9039-0.9962)} & (0.9440-1.000) \\ \addlinespace[0.4em]
    Kappa & 0.8723 & 0.8434 & 0.8577 & 0.7738 & 0.7842 & 0.8431 & \textbf{0.9136} \\ \addlinespace[0.4em]
    AUC   & 0.9895 & 0.9877 & 0.9796 & 0.9609 & 0.9707 & 0.9706 & \textbf{0.9948} \\ \addlinespace[0.4em]
    AP    & 0.9833 & 0.9798 & 0.9686 & 0.9428 & 0.9504 & 0.9588 & \textbf{0.9905} \\
    \bottomrule
    \end{tabular}%
\end{center}
\end{table*}

Then, Table~\ref{table4} summarises the overall performance of all AI models, indicating that our model yields the best results in all metrics. Detailed results across categories are provided in Supplementary Table~\ref{sup:tab8}. According to the confusion matrix of our model (Supplementary Fig.~\ref{sup:fig4}), 106 of 122 patients are correctly diagnosed by our model, which is the highest among the AI models. Importantly, all patients with EGJA are not missed, and only two patients with E-EGJA are over-diagnosed by our model as A-EGJA. In contrast, other AI models suffer from the miss-diagnosis of EGJA into control or the under-staging of A-EGJA into E-EGJA.

Next, to validate the effect of demographic variables on the performance of our model, we group these 112 patients by sex (male/female) or age (age<60, 60$\leq$ age<70, age$\geq$70). Supplementary Table~\ref{sup:tab9} shows the results of our model in different patient groups. We observe that our model exhibits comparable classification accuracy across different age groups (0.9535 for patients aged <60 years, 0.9412 for patients aged 60-70 years; 0.9429 for patients aged$\geq$70 years). In contrast, while our model maintains satisfactory performance in both sexes, a higher accuracy is observed for male patients compared to female patients (0.9595 vs. 0.9211). This discrepancy in sex may be attributed to the significantly higher incidence of EGJA in males (the male-to-female incidence rate ratio is about 4:1 \cite{ref43}), resulting into 330 female patients vs. 676 male patients in our training set.

\subsection{Comparison to endoscopists with varying experience levels}

To further evaluate the effectiveness of our model, we compare its classification results against endoscopists with varying experience levels on the held-out test set at the image level. Table~\ref{table5} presents the classification results for each endoscopist group on the held-out test set while Table~\ref{table6} summarises results across all categories. The classification results for each endoscopist are detailed in Supplementary Table~\ref{sup:tab10}-\ref{sup:tab12}.

\begin{table*}[!t]
\caption{{Comparison of three endoscopist groups A without the assistance of our model and three endoscopist groups B with the assistance of our model on the held-out test set at the image level.}}
%\vspace{-0.1in}
\label{table5}
\begin{center}
\scriptsize
\setlength{\tabcolsep}{1.5mm}
    \begin{tabular}{cccccccc}
    \toprule
    \multirow{2}{*}{} & \textbf{Trainee} & \textbf{Competent} & \textbf{Expert} & \textbf{Trainee} & \textbf{Competent} & \textbf{Expert} & \multirow{2}{*}{\textbf{Our model}} \\
          & \textbf{group A} & \textbf{group A} & \textbf{group A} & \textbf{group B} & \textbf{group B} & \textbf{group B} &  \\
    \midrule
    {Accuracy} & 0.7035 & 0.735 & 0.8147 & 0.8497 & 0.8521 & 0.8696 & \textbf{0.9256} \\
    {(95\% CI)} & (0.6739-0.7331) & (0.7064-0.7636) & (0.7895-0.8399) & (0.8265-0.8728) & (0.8291-0.8751) & (0.8478-0.8914) & (0.9086-0.9426) \\ \addlinespace[0.4em]
    {Sensitivity} & 0.7552 & 0.7668 & 0.8231 & 0.8528 & 0.857 & 0.8708 & \textbf{0.9227} \\
    {(95\% CI)} & (0.7092-0.8012) & (0.7181-0.8156) & (0.7806-0.8656) & (0.8118-0.8938) & (0.8195-0.8944) & (0.8316-0.9099) & (0.8946-0.9509) \\ \addlinespace[0.4em]
    {Specificity} & 0.8666 & 0.8782 & 0.9077 & 0.9232 & 0.9254 & 0.933 & \textbf{0.9581} \\
    {(95\% CI)} & (0.8418-0.8914) & (0.8542-0.9022) & (0.8842-0.9311) & (0.9018-0.9446) & (0.9040-0.9468) & (0.9130-0.9531) & (0.9423-0.9738) \\ \addlinespace[0.4em]
   {PPV} & 0.7057 & 0.7374 & 0.7864 & 0.8281 & 0.8266 & 0.85  & \textbf{0.9189} \\
    {(95\% CI)} & (0.6611-0.7502) & (0.6931-0.7818) & (0.7401-0.8327) & (0.7854-0.8708) & (0.7828-0.8705) & (0.8093-0.8907) & (0.8877-0.9501) \\ \addlinespace[0.4em]
    {NPV} & 0.8502 & 0.8606 & 0.8965 & 0.9142 & 0.9163 & 0.9253 & \textbf{0.957} \\
    {(95\% CI)} & (0.8260-0.8744) & (0.8355-0.8857) & (0.8741-0.9189) & (0.8929-0.9355) & (0.8962-0.9365) & (0.9050-0.9455) & (0.9419-0.9721) \\ \addlinespace[0.4em]
    Kappa & 0.554 & 0.5922 & 0.7022 & 0.7554 & 0.7599 & 0.7869 & \textbf{0.8763} \\ \addlinespace[0.4em]
    P value & <0.0001 & <0.0001 & 0.0003 & <0.0001 & 0.0006 & <0.0001 & Reference \\\addlinespace[0.4em]
    Time cost(s) & 4.209 & 4.261 & 2.795 & 4.611 & 5.309 & 5.296 & \textbf{0.0158} \\
    \bottomrule
    \end{tabular}%
\end{center}
\end{table*}

\begin{table*}[!t]
\caption{{Comparison of three endoscopist groups A without the assistance of our model and three endoscopist groups B with the assistance of our model across categories on the held-out test set at the image level.}}
%\vspace{-0.1in}
\label{table6}
\begin{center}
\scriptsize
\setlength{\tabcolsep}{1.5mm}
    \begin{tabular}{cccccccc}
    \toprule
    \multirow{2}{*}{} & \textbf{Trainee} & \textbf{Competent} & \textbf{Expert} & \textbf{Trainee} & \textbf{Competent} & \textbf{Expert} & \multirow{2}{*}{\textbf{Our model}} \\
          & \textbf{group A} & \textbf{group A} & \textbf{group A} & \textbf{group B} & \textbf{group B} & \textbf{group B} &  \\
    \midrule
    \multicolumn{8}{c}{\textbf{Classification results on the A-EGJA category}} \\
    \multicolumn{1}{c}{Accuracy} & 0.7597 & 0.7937 & 0.8479 & 0.8746 & 0.8777 & 0.8919 & \textbf{0.93} \\
    \multicolumn{1}{c}{(95\% CI)} & (0.7320-0.7874) & (0.7674-0.7956) & (0.8246-0.8712) & (0.8531-0.8961) & (0.8564-0.8989) & (0.8718-0.9120) & (0.9134-0.9465) \\ \addlinespace[0.4em]
    \multicolumn{1}{c}{Sensitivity} & 0.5911 & 0.666 & 0.796 & 0.8427 & 0.841 & 0.8668 & \textbf{0.9316} \\
    \multicolumn{1}{c}{(95\% CI)} & (0.5479-0.6344) & (0.6245-0.7075) & (0.7605-0.8314) & (0.8106-0.8747) & (0.8089-0.8732) & (0.8369-0.8967) & (0.9094-0.9538) \\ \addlinespace[0.4em]
    \multicolumn{1}{c}{Specificity} & \textbf{0.9607} & 0.9458 & 0.9098 & 0.9127 & 0.9213 & 0.9218 & 0.9281 \\
    \multicolumn{1}{c}{(95\% CI)} & (0.9420-0.9793) & (0.9241-0.9675) & (0.8823-0.9373) & (0.8856-0.9398) & (0.8955-0.9472) & (0.8961-0.9476) & (0.9033-0.9529) \\ \addlinespace[0.4em]
    \multicolumn{1}{c}{PPV} & \textbf{0.9471} & 0.9361 & 0.9132 & 0.92  & 0.9272 & 0.9297 & 0.9391 \\
    \multicolumn{1}{c}{(95\% CI)} & (0.9222-0.9720) & (0.9106-0.9616) & (0.8867-0.9397) & (0.8951-0.9450) & (0.9033-0.9512) & (0.9064-0.9529) & (0.9180-0.9603) \\ \addlinespace[0.4em]
    NPV   & 0.6635 & 0.7038 & 0.7891 & 0.8296 & 0.8294 & 0.8531 & \textbf{0.9192} \\
    (95\% CI) & (0.6258-0.7012) & (0.6660-0.7416) & (0.7526-0.8256) & (0.7951-0.8640) & (0.7952-0.8637) & (0.8204-0.8858) & (0.8932-0.9453) \\ \addlinespace[0.4em]
    P value & <0.0001 & <0.0001 & <0.0001 & <0.0001 & <0.0001 & <0.0001 & Reference \\ \midrule
    \multicolumn{8}{c}{\textbf{Classification results on the E-EGJA category}} \\
    Accuracy & 0.7617 & 0.7683 & 0.8573 & 0.8744 & 0.8864 & 0.8908 & \textbf{0.9311} \\
    (95\% CI) & (0.7342-0.7893) & (0.7409-0.7956) & (0.8347-0.8800) & (0.8529-0.8959) & (0.8659-0.9070) & (0.8706-0.9111) & (0.9146-0.9475) \\ \addlinespace[0.4em]
    Sensitivity & 0.75  & 0.8048 & 0.7327 & 0.7904 & 0.7567 & 0.8183 & \textbf{0.8462} \\
    (95\% CI) & (0.6912-0.8088) & (0.7509-0.8587) & (0.6725-0.7928) & (0.7351-0.8457) & (0.6984-0.8150) & (0.7659-0.8707) & (0.7971-0.8952) \\ \addlinespace[0.4em]
    Specificity & 0.7652 & 0.7575 & 0.8941 & 0.8992 & 0.9246 & 0.9122 & \textbf{0.9561} \\
    (95\% CI) & (0.7339-0.7964) & (0.7259-0.7891) & (0.8713-0.9168) & (0.8769-0.9214) & (0.9052-0.9441) & (0.8913-0.9331) & (0.9410-0.9712) \\ \addlinespace[0.4em]
    PPV   & 0.4848 & 0.4944 & 0.6708 & 0.6978 & 0.7474 & 0.733 & \textbf{0.8502} \\
    (95\% CI) & (0.4302-0.5394) & (0.4411-0.5476) & (0.6097-0.7319) & (0.6392-0.7564) & (0.6887-0.8061) & (0.6761-0.7899) & (0.8016-0.8989) \\ \addlinespace[0.4em]
    NPV   & 0.9122 & 0.9294 & 0.919 & 0.9357 & 0.9281 & 0.9446 & \textbf{0.9547} \\
    (95\% CI) & (0.8894-0.9350) & (0.9085-0.9504) & (0.8986-0.9394) & (0.9173-0.9542) & (0.9090-0.9472) & (0.9274-0.9617) & (0.9394-0.9701) \\ \addlinespace[0.4em]
    P value & <0.0001 & <0.0001 & 0.0002 & <0.0001 & 0.0056 & <0.0001 & Reference \\ \midrule
    \multicolumn{8}{c}{\textbf{Classification results on the control category}} \\
    Accuracy & 0.8856 & 0.9081 & 0.9241 & 0.9503 & 0.94  & 0.9565 & \textbf{0.9902} \\
    (95\% CI) & (0.8649-0.9062) & (0.8894-0.9268) & (0.9069-0.9412) & (0.9362-0.9644) & (0.9247-0.9554) & (0.9432-0.9697) & (0.9838-0.9966) \\ \addlinespace[0.4em]
    Sensitivity & 0.9244 & 0.8297 & 0.9407 & 0.9254 & 0.9732 & 0.9273 & \textbf{0.9904} \\
    (95\% CI) & (0.8886-0.9602) & (0.7787-0.8806) & (0.9086-0.9727) & (0.8897-0.9610) & (0.9513-0.9951) & (0.8921-0.9625) & (0.9772-1.0000) \\ \addlinespace[0.4em]
    Specificity & 0.874 & 0.9313 & 0.9191 & 0.9577 & 0.9302 & 0.9651 & \textbf{0.9901} \\
    (95\% CI) & (0.8495-0.8985) & (0.9127-0.9500) & (0.8990-0.9393) & (0.9429-0.9726) & (0.9114-0.9490) & (0.9516-0.9787) & (0.9828-0.9974) \\ \addlinespace[0.4em]
    PPV   & 0.6851 & 0.7818 & 0.7752 & 0.8665 & 0.8052 & 0.8874 & \textbf{0.9673} \\
    (95\% CI) & (0.6309-0.7393) & (0.7274-0.8361) & (0.7239-0.8266) & (0.8219-0.9111) & (0.7564-0.8541) & (0.8454-0.9293) & (0.9435-0.9911) \\ \addlinespace[0.4em]
    NPV   & 0.975 & 0.9486 & 0.9812 & 0.9774 & 0.9915 & 0.9781 & \textbf{0.9971} \\
    (95\% CI) & (0.9628-0.9872) & (0.9321-0.9650) & (0.9709-0.9916) & (0.9663-0.9885) & (0.9846-0.9985) & (0.9673-0.9890) & (0.9932-1.0000) \\ \addlinespace[0.4em]
    P value & <0.0001 & 0.0021 & <0.0001 & <0.0001 & <0.001 & 0.0011 & Reference \\
    \bottomrule
    \end{tabular}%
\end{center}
\end{table*}

According to Table~\ref{table5}, our model significantly outperforms each endoscopist group A in all metrics (P<0.05, the McNemar-Bowker test). Particularly, our model exhibits a substantial advantage over expert group A in terms of time cost (0.0158s vs. 2.795s). These results indicate the considerable potential of our model to enhance diagnostic efficiency and accuracy. Furthermore, the superiority of our model over endoscopists is also evident across categories (Table~\ref{table6}). For example, in the E-EGJA category, the highest specificity (0.9561) and NPV (0.9547) of our model show its reliability in distinguishing patients without E-EGJA, whereas all endoscopist groups suffer from significant deficiencies in certain metrics (e.g., PPV being 0.4848 for trainee group A, 0.4944 for competent group A, and 0.6708 for expert group A).

\begin{figure*}[!t]
\begin{center}
\begin{tabular}{c}
\includegraphics[width=4.3in]{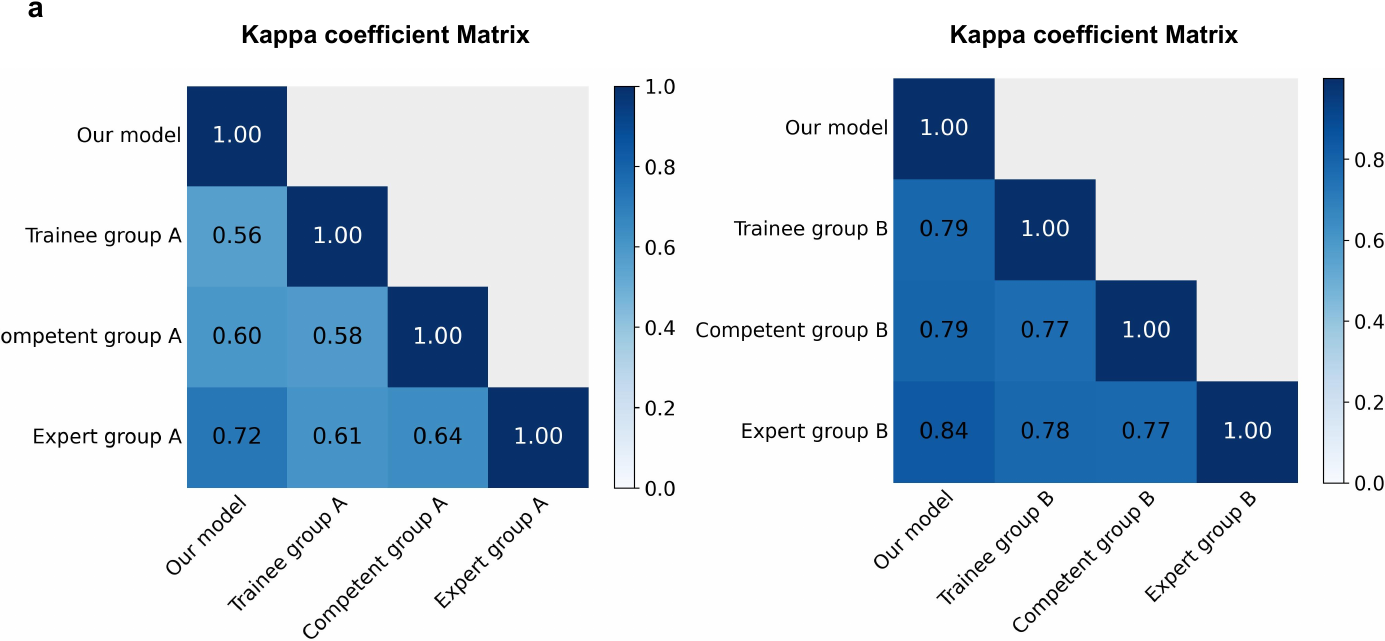} \\
\includegraphics[width=5.5in]{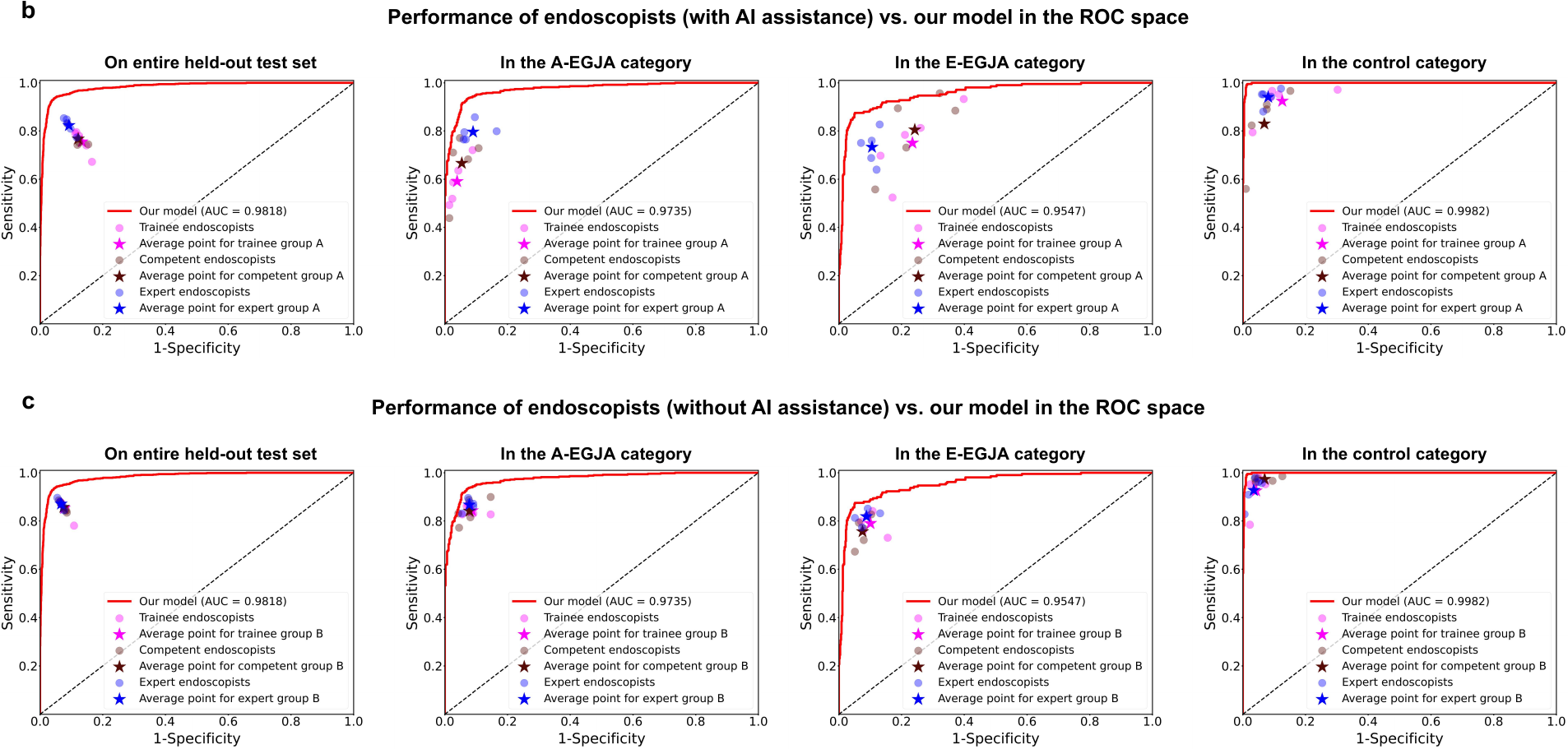}  \\
\includegraphics[width=5.5in]{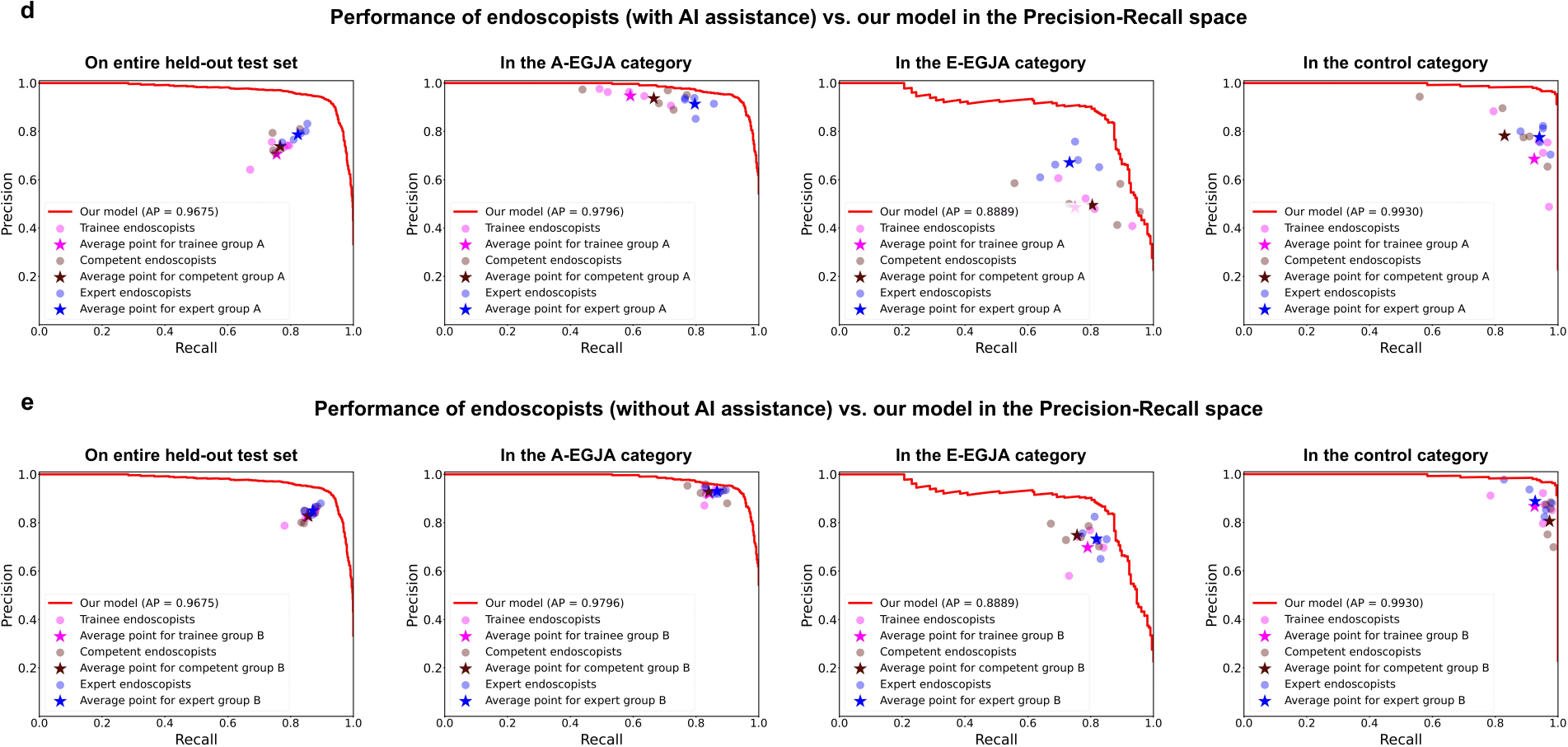} \\
\end{tabular}
\end{center}
\setlength{\abovecaptionskip}{-5pt}
\caption{Performance of different endoscopist groups vs. our model on the held-out test set. \textbf{a:} degree of diagnostic consistency between our model and different endoscopist groups.\textbf{b:} the performance of each group A without AI assistance in the ROC space. \textbf{c:} the performance of each group B with AI assistance in the ROC space. \textbf{d:} the performance of each group A without AI assistance in the precision-recall space. \textbf{e:} the performance of each group B with AI assistance in the precision-recall space. The circles in the same colour depict specificity and sensitivity of 5 endoscopists in each group while the star represents the average point of 5 endoscopists in each group.}
\label{fig3}
\end{figure*}

Next, the comparison is visualised by confusion matrices (Supplementary Fig.~\ref{sup:fig5}a). We observe that expert group A performs the best among the three groups, but is still considerably worse than ours (Fig.~\ref{fig2}a). Fig.~\ref{fig3}a shows the diagnostic consistency (calculated by the Kappa consistency test) between our model and different endoscopist groups on the held-out test set. We can observe from Fig.~\ref{fig3}a left that, the diagnostic consistency, denoted by the Kappa coefficient, between our model and three endoscopist groups is 0.56, 0.60, and 0.72, respectively. Fig.~\ref{fig3}b and d visualise the performance of each group A by plotting the points (sensitivity and specificity, precision and recall) of endoscopists in the ROC and precision-recall spaces, respectively. We observe that the average point of 5 endoscopists in each group, denoted by the star in Fig.~\ref{fig3}b and d, is consistently situated below the ROC and precision-recall curves of our model. Moreover, the individual points for 15 endoscopists in the three groups, denoted by circles in Fig.~\ref{fig3}b and d, reveal considerable variation in their EGJA staging diagnosis, even among those with similar experience levels (see circles in the same colour). As a result, different endoscopists can be inconsistent in their diagnosis of the same patient, which may incur additional medical burden to the patient and even lead to severe consequences.

\subsection{Endoscopists’ performance with the assistance of our model}

To assess our model for the AI-assisted diagnostic application, we conduct an AI-assisted evaluation on the held-out test set at the image level. As shown in Table~\ref{table5}, with the assistance of our model, each endoscopist group B yields sizeable improvements over corresponding group A. Particularly, trainee group B obtains an accuracy improvement of +14.6\% compared with trainee group A while competent group B obtains an accuracy improvement of +11.7\% compared with competent group A. Consequently, both groups B can yield competitive performance on most metrics compared with expert group B (e.g., accuracy being 0.8497 for trainee group B, 0.8521 for competent group B, and 0.8696 for expert group B). 

Next, as probed from Supplementary Fig.~\ref{sup:fig5}a, trainee group A and competent group A have the tendency to misdiagnose A-EGJA as E-EGJA; while with the assistance of our model, the occurrence of this error is significantly reduced (Supplementary Fig.~\ref{sup:fig5}b). Then, as shown in Fig.~\ref{fig3}a right, the diagnostic consistency between our model and three endoscopist groups is clearly increased to 0.79, 0.79, and 0.84, respectively. Meanwhile, the consistency between any two endoscopist groups is also increased from 0.58, 0.61, 0.64 to 0.77, 0.78, 0.77, respectively. In Fig.~\ref{fig3}c and e, we compare the performance of each group B in the ROC and precision-recall spaces with our model. As depicted in Fig.~\ref{fig3}c and e, the points representing 5 endoscopists in every group B are more tightly clustered in both ROC and precision-recall spaces compared with those in every group A (Fig.~\ref{fig3}b and d). Meanwhile, the distances from the average points to the curves are significantly shorter for groups B than for groups A, respectively. These findings suggest that our model can enhance endoscopists’ performance while reducing their variations by assisting them to achieve better diagnostic accuracy.

\subsection{Model visualisation and interpretation}

\begin{figure*}[!t]
  \centerline{\includegraphics[width=6in]{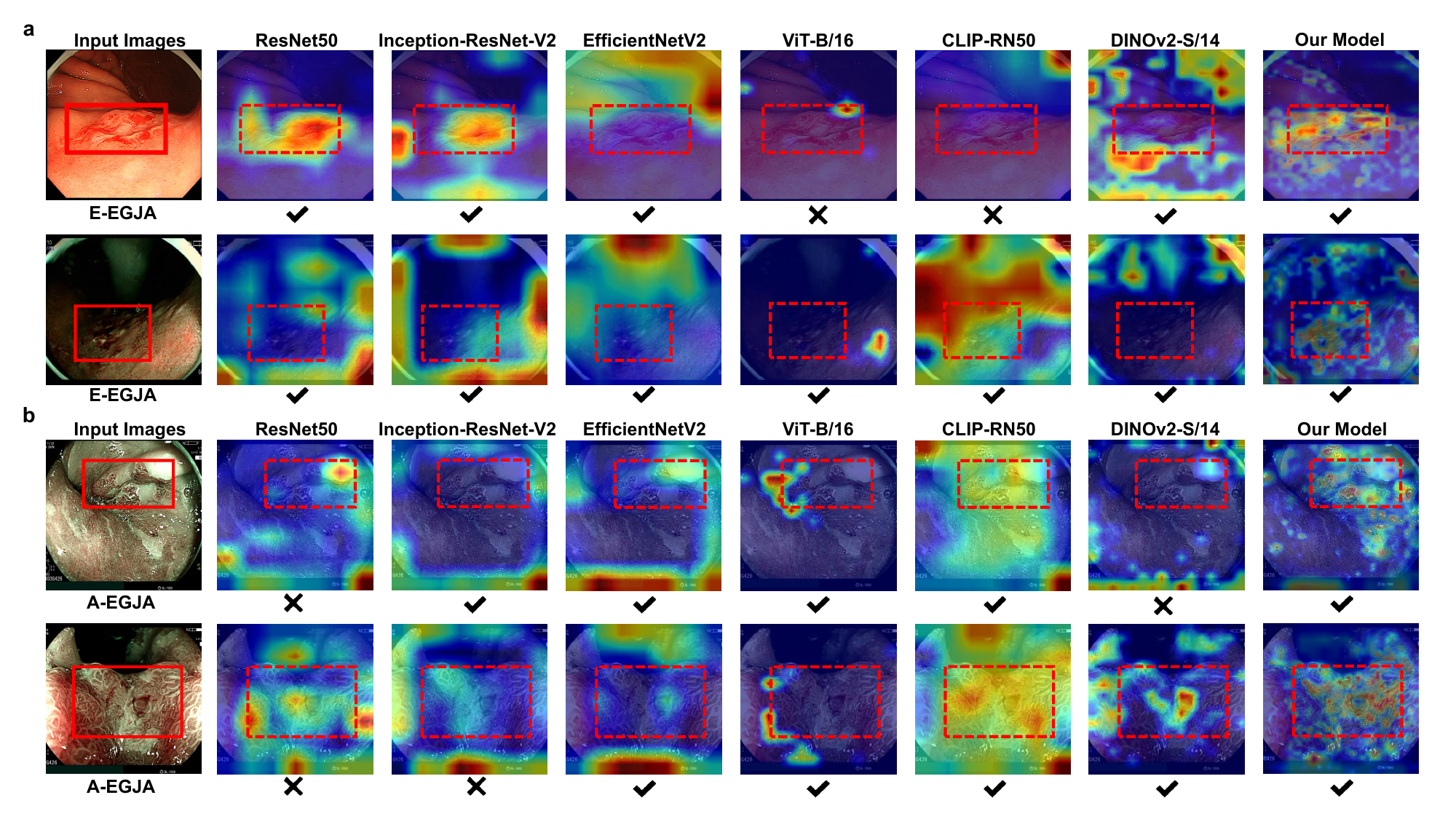}}
  \caption{Comparison of the Grad-CAMs by each AI model in diagnosing the E-EGJA \textbf{(a)} and A-EGJA \textbf{(b)} on the held-out test set. The colours from blue to red in each Grad-CAM denote the activation values from low to high, depending on which the model makes the prediction. The red bounding boxes in input images are annotated by chief endoscopists.}
  \label{fig4}
\end{figure*}

We visualise the regions of interest for all AI models in diagnosing E-EGJA and A-EGJA by the Grad-CAMs (Fig.~\ref{fig4}, Supplementary Fig.~\ref{sup:fig7} and \ref{sup:fig8}). As shown in the figures, the colours from blue to red denote the activation values from low to high, depending on which the AI model makes the prediction, and the activation area indicated by our model matches the best with the annotated ground truth area (the red bounding box).

\begin{figure*}[!t]
  \centerline{\includegraphics[width=6in]{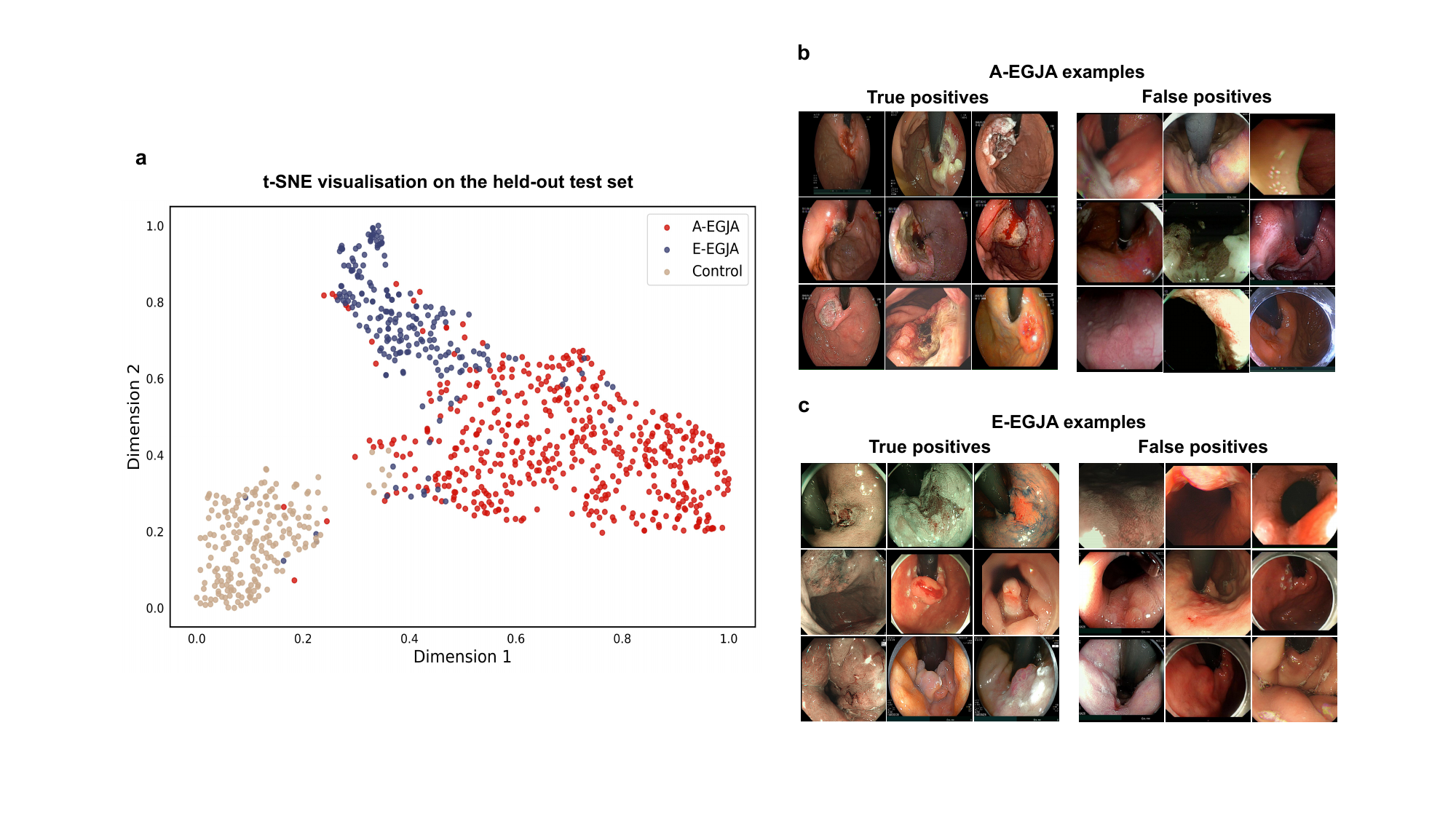}}
  \caption{The t-SNE visualisation of our model (left) and image examples diagnosed by our model as A-EGJA and E-EGJA on the held-out test set (right). \textbf{a:} the two-dimensional t-SNE visualisation of our model. Each point represents the two-dimensional embeddings of a single image in the test set. \textbf{b:} A-EGJA examples diagnosed by our model. \textbf{c:} E-EGJA examples diagnosed by our model. }
  \label{fig5}
\end{figure*}

As shown in the EGJA images (Supplementary Fig.~\ref{sup:fig7} and \ref{sup:fig8}), E-EGJA lesions (the red bounding box) appear as slightly morphologically irregular, small depressed or elevated lesions. Clear boundaries and colour tone changes can be observed in the images, with the surrounding mucosa being basically flat and continuous. In contrast, A-EGJA lesions (the red bounding box) are characterised by large elevated-depressed or ulcerative lesions. The surface of A-EGJA shows abundant dirty exudate accompanied by spontaneous bleeding, and the surrounding mucosa is often associated with reactive thickening or interruption of folds. These distinct features between E-EGJA and A-EGJA can also be viewed in Fig.~\ref{fig5}b and c, Supplementary Fig.~\ref{sup:fig9}b and c, and Supplementary Fig.~\ref{sup:fig10}b and c, in which we show some representative examples (true positives) of E-EGJA and A-EGJA diagnosed by our model on the three test sets.

Next, in Fig.~\ref{fig4}, we choose those images that are misdiagnosed by the endoscopists from expert group A. Specifically, in Fig.~\ref{fig4}a, the first column shows two E-EGJA images. For the first image, all 5 endoscopists misdiagnose it as A-EGJA. For the second image, 3 endoscopists make the right diagnosis while 2 misdiagnose it as A-EGJA. We observe that the E-EGJA lesions in these two images appear as a non-protruding depressed type on the posterior wall. For the two A-EGJA images from the first column in Fig.~\ref{fig4}b, 4 endoscopists misdiagnose them as E-EGJA while only one diagnoses them correctly. We observe that the A-EGJA lesions in these two images are present as a flat and depressed type at the squamocolumnar junction. These findings indicate that for certain types of E-EGJA and A-EGJA lesions, their visual features could be similar, making them difficult to distinguish by endoscopists.

Furthermore, Fig.~\ref{fig5}a, Supplementary Fig.~\ref{sup:fig9}a, and \ref{sup:fig10}a present the t-SNE visualisation of our model on the three test sets. We observe the principal separation between A-EGJA and E-EGJA images in the t-SNE space, except for the mixed points that form an overlapping region. To understand why our model fails to correctly diagnose the EGJA stages, two chief endoscopists evaluate the misdiagnosed images (shown as false positives in Fig.~\ref{fig5}, Supplementary Fig.~\ref{sup:fig9} and \ref{sup:fig10}) of our model on the three test sets. The reasons are summarised as: (1) Insufficient lesion exposure due to under-inflation of the gastrointestinal tract or restricted visualisation; (2) Atypical and subtle morphological features in non-protruding depressed lesions.
\section{Discussion}

This study reports the development of an AI model for EGJA staging diagnosis by leveraging the vision foundation model and CNN. The developed model has achieved robust and superior performance to other AI models and experienced endoscopists on the test sets. The AI-assisted evaluation in this study further attests to the effectiveness of our model as an auxiliary tool for EGJA diagnosis, while the visual analysis offers certain interpretability of the AI results, underscoring the potential clinical benefits.

In recent years, computer-aided image analysis technology has made significant progress in enhancing the diagnostic accuracy of upper gastrointestinal diseases \cite{ref44}. According to a recent systematic review \cite{ref45}, current AI models have shown strong performance in diagnosing gastric cancers. Some of them can be directly used for the EGJA diagnosis \cite{ref32,ref46,ref47}. For instance, Luo et al.\cite{ref32} reported the GRAIDS system developed from over a million endoscopic images, achieving high diagnostic rates for most upper gastrointestinal cancers. However, due to the relatively low incidence of EGJA (3.5 per 10000 \cite{ref48}), the number of EGJA cases is inadequate to achieve a good outcome. Moreover, the effectiveness of the staging diagnosis of EGJA (i.e., E-EGJA and A-EGJA) remains to be assessed. In a recent study on the E-EGJA diagnosis \cite{ref17}, the AI diagnostic model is developed based on Japanese data including 202 pathologically confirmed E-EGJA cases and 262 cases with normal mucosa. However, considering the low incidence of EGJA in Japan and its single-centre design, the developed AI model is not satisfactory, only achieving an accuracy of 66\% for E-EGJA while incapable of the staging diagnosis. Overall, the limited sample size, the lack of multicentre data, and the absence of staging diagnosis so far pose substantial challenges to develop the AI model for robust EGJA staging diagnosis.

To address these issues, this study conducts multicentric data collection. Specifically, 12,302 endoscopic images from 1,546 patients are retrospectively collected from seven hospitals. As a result, a multicentric EGJA dataset is constructed for our model development. On this basis, our model can be fully trained for EGJA diagnosis as well as comparative evaluations. Compared with existing AI models for EGJA diagnosis, the main advantage of our model lies in its ability to perform EGJA staging diagnosis while achieving a very good diagnostic rate for E-EGJA (0.9311, 0.9012, and 0.9125 on the held-out, external, and prospective test set, respectively). Besides, to the best of our knowledge, this is the largest study in Asia conducted for AI-guided EGJA diagnosis from single endoscopic image.

Many previous studies directly applied classical CNNs to achieve endoscopic diagnosis of specific diseases without making significant adjustments \cite{ref11,ref12,ref13,ref14,ref15,ref17}. However, the quantitative performance of their models largely depends on the volume of their training sets, which may lead to incompatibility for EGJA staging diagnosis. In contrast, our model has exploited state-of-the-art AI techniques: we introduce an advanced vision foundation model, the pretrained DINOv2, to serve as an encoder in our model. It has a strong generalisability to learn robust representations from endoscopic images. In addition, the utilisation of MoE has enabled our model to achieve elaborate feature fusion of global appearance and local details of input images. These techniques have considerably improved the performance of our model on all quantitative results compared with models built solely on CNNs (i.e., ResNet50, Inception-ResNet-V2, and EfficientNetV2) or vision foundation models (i.e., ViT-B/16, CLIP-RN50, and DINOv2-S/14). The superiority of our model is further attested by its robust performance on both external and prospective test sets. 

Next, we compare our results against those performed by endoscopist groups with varying experience levels, demonstrating that our model can outperform all endoscopists recruited in this study. In particular, the PPV (0.8502) of our model for E-EGJA diagnosis is significantly higher than that of endoscopists (0.4848, 0.4944, and 0.6708), reflecting the reliability of our model in diagnosing E-EGJA. In contrast, endoscopists generally suffer from EGJA staging diagnosis. The primary reason for this is that the conclusive and widely accepted endoscopic diagnostic criteria for E-EGJA have not yet been established. The diagnosis of E-EGJA is usually made by endoscopists referring to the early endoscopic manifestations of esophageal or gastric cancer or cancer. However, the malignancy grade of E-EGJA lesions under endoscopy tends to be more severe than that of esophageal or gastric cancer. One meta-analysis reveals that over 20\% of patients with E-EGJA exhibit deep submucosal invasion in pathology findings following endoscopic submucosal dissection \cite{ref49}, which implies that the accuracy of endoscopists’ diagnosis of E-EGJA still requires further enhancement. Therefore, it is insufficient to simply make staging judgments for EGJA based on the criteria for gastric or esophageal cancer. 

The challenges that endoscopists face in accurately diagnosing E-EGJA are multifaceted. For instance, under endoscopy, asymptomatic presentations and subtle mucosal changes in E-EGJA lesions contribute significantly to the difficulty in distinguishing them from normal tissues. This diagnostic dilemma is further complicated by the fact that E-EGJA is often flat and subtle, with the anatomy of the esophagogastric junction, increasing the difficulty of detection. As a result, this diagnosis is subjective and to a great extent relies on the proficiency and experience of the endoscopists. To address it, several advanced techniques can be leveraged, such as special staining marking and magnifying endoscopy, which have shown promising potential in differentiating between cancerous and non-cancerous lesions. However, their clinical applicability has been jeopardised because of the drawbacks of high cost, poor universality, and the long process for training qualified operators.

Besides, according to the findings in our visual analysis, despite exhibiting distinguishable features, E-EGJA and A-EGJA also demonstrate overlapping features in certain cases. For instance, E-EGJA with non-protruding depressed-type lesions is associated with a heightened likelihood of submucosal invasion, potentially leading endoscopists to misdiagnose it as A-EGJA. In contrast, A-EGJA with flat and depressed lesions at the squamocolumnar junction is often elusive under endoscopic examination, which may be misdiagnosed as E-EGJA by endoscopists. Based on these findings, when EGJA lesions display a raised-depressed morphology, and are located near the squamocolumnar junction, a more cautious assessment of their staging should be considered by endoscopists.

In contrast to the endoscopists, our model can be learned at a low cost yet exhibits high precision and good generalisability. However, the superiority of our AI model does not mean that it can replace endoscopists for EGJA diagnosis in clinical practice. In fact, the affirmative clinical diagnosis of EGJA requires more than just the lesion characteristics shown in endoscopic images, but also the clues from the clinical context and the integration of several ancillary methods such as pathological assessments and biochemical tests. The AI model, on the other hand, demonstrates how to provide valuable and reliable support to make the final decision. 

For this purpose, we design an AI-assisted evaluation to demonstrate the feasibility of our model in clinical diagnostic application. With the assistance of our model, the overall performance of the trainee, competent, and expert groups has improved significantly. Importantly, our AI model can identify the subtle morphological patterns of E-EGJA, reducing the false-negative rates of endoscopists for E-EGJA diagnosis (e.g., on the held-out test set, the trainee endoscopists initially missed 25\% of E-EGJA images, but with our model’s assistance, their missing rate was reduced to 15\%). While A-EGJA may have obvious surface features, the model’s value lies in differentiating resectable T1b lesions from non-resectable T2 cases as well as guiding treatment decisions for borderline cases. In particular, with the assistance of our model, the sensitivity of trainee and competent endoscopists for A-EGJA diagnosis can be significantly improved from 0.5911 and 0.6660 to 0.7960 and 0.8427, respectively. Therefore, our model can provide certain support for less experienced endoscopists, enhancing their diagnostic rate while addressing the shortage of expert clinicians.

Lastly, this study still has several limitations. First, the retrospective nature of our dataset may introduce some biases, such as demographic bias and survivorship bias. Second, our model is trained exclusively on white-light and NBI images, as they are commonly used in clinical practice. Consequently, our model has not been evaluated against other endoscopic modalities. Third, our data is primarily collected from hospitals in Shanghai, which may not cover all lesion characteristics of EGJA, hence, lacking prospective validation over diverse populations. Fourth, the development of our model lacks integration with multimodal clinical data (e.g., histopathology and genetic markers). In the future, we will make efforts to collect endoscopic images of other modalities (e.g., magnified endoscopic images) while expanding the coverage of our data to diverse populations. Meanwhile, we will also focus on integrating multimodal data into our model development and exploring real-time deployment in clinical workflows. Additionally, exploring advanced deep learning techniques to enhance the generalisability of our model could also be a valuable direction for future research.

\section*{Contributors}
Y.M. constructed the model. Y.M., J.L., L.M., and Z.Y. optimised the model. M.S. presided over the construction and optimisation of the model. Y.C. and S.X. designed the multicentre study. B.L. was responsible for data collection. L.Z., D.Z., L.X., Y.Z., X.L., W.D., M.Z., D.H., Z.L., Y.C., Y.Z., J.Z., X.W., and L.Y. participated in the setup standardisation, data collection, and image labelling. H.S. participated in the whole process of collaborative research. Y. M. and B.L. drafted the manuscript. Y.C., M.S., S.X., H.S., and Z.Y. revised the manuscript. M.S. and H.S. have accessed and verified the underlying study data. All authors read and approved the final version of the manuscript.

\section*{Data sharing}
The dataset utilised in this study is currently restricted from public release due to data privacy laws and the policies of the respective institutional review boards. If the interested researchers want to obtain the dataset for non-commercial use, they can request for the corresponding authors (M.S. or H.S.). Corresponding authors will review their requests and ask for consent from each centre.

\section*{Declaration of interests}
All authors declare no competing interests.

\section*{Acknowledgments}
This study was supported by the following funding sources: Clinical Research Plan of Shanghai Health Development Commission (grant No.SHDC2022CRT004); Science and Technology Innovation Action Plan of Shanghai Science and Technology Commission (grant No.22DZ2203900); Independent and Original Basic Research of Tongji University (grant No.15082150043); “Medical+X” Interdisciplinary Research Program of Tongji University (grant No.2025-0313-ZD-01); The 19th Experimental Teaching Reform Project of Tongji University (grant No.1508104012); Shanghai Science and Technology Commission Project (grant No.21Y11908500); Shanghai Science and Technology Innovation Action Plan (Project for Popular Science); Shanghai Municipal Commission of Economy and Informatization Project (Demonstration Project for the Application of Innovative Medical Devices, grant No.23SHS03100, No.23SHS03100-01, No.23SHS03100-03). The authors would like to thank all the doctors who participated in the collection and processing of the images and the subsequent AI-assisted evaluation in this study.

\bibliographystyle{elsarticle-num-names} 
\bibliography{Mycollection.bib}

\clearpage

\noindent
\begin{center}
  \fontsize{16}{20}\selectfont % 设置字体大小和行距
  \bfseries % 加粗
  Supplementary Material
\end{center}

% 添加一些垂直间距
% \vspace{1cm}

% 重置图表计数器并使用S编号
\setcounter{figure}{0}
\renewcommand{\thefigure}{S\arabic{figure}}
\setcounter{table}{0}
\renewcommand{\thetable}{S\arabic{table}}

% 为补充材料中的图表创建新标签前缀
\makeatletter
\newcommand{\suplabel}[1]{\label{sup:#1}}
\newcommand{\supref}[1]{Supplementary~Figure~\ref{sup:#1}}
\makeatother

\section*{The description of our model}

\begin{figure}[htbp]
\centering
\includegraphics[width=0.95\textwidth]{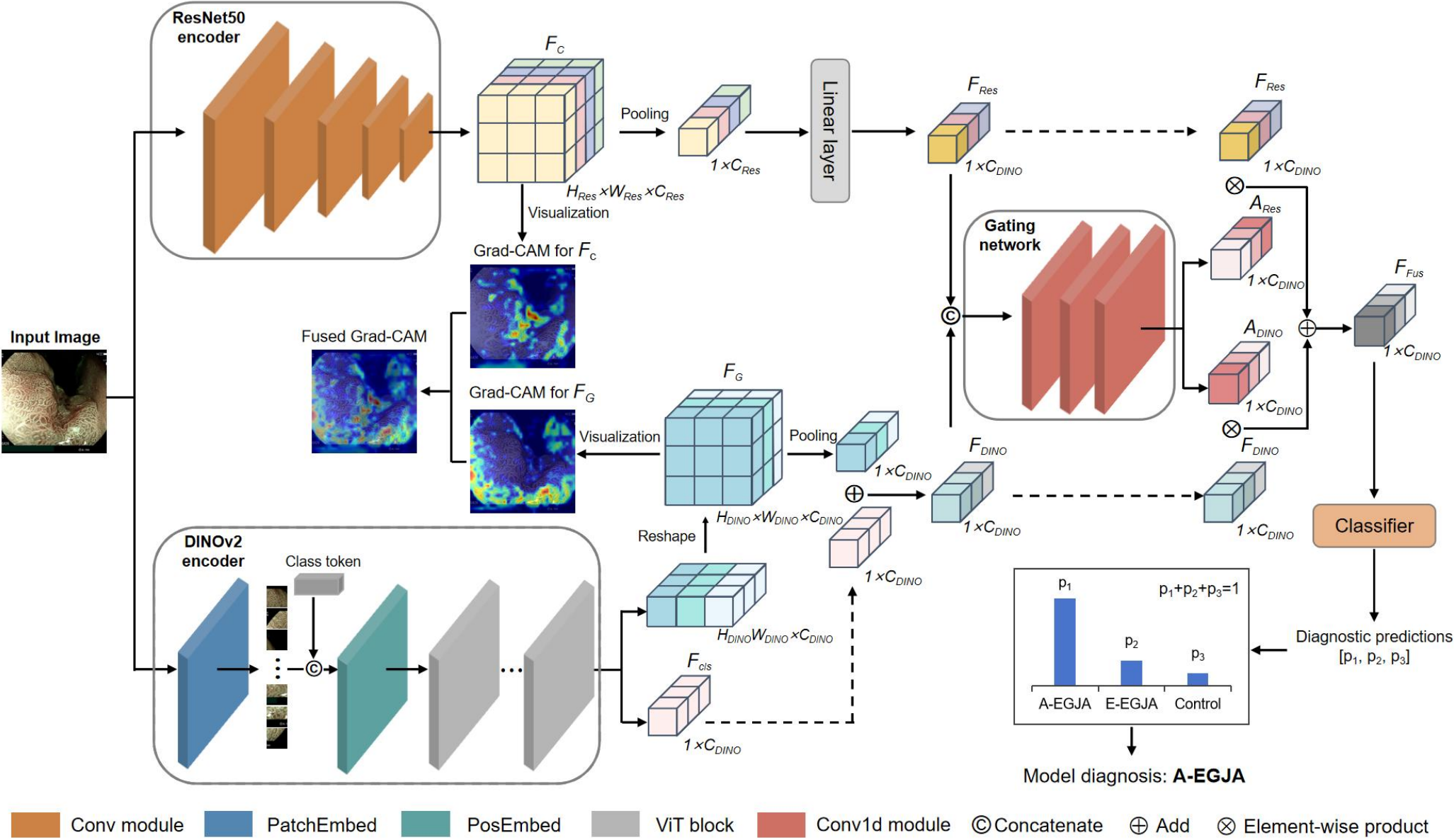}

\caption{Framework of our AI diagnostic model for EGJA. It mainly consists of two image encoders, a gating network and a classifier. Features extracted by the ResNet50 and DINOv2 encoders are denoted by $F_C$, $F_G$, and $F_{cls}$. $F_C$ and $F_G$ can be visualised by the gradient-weighted class activation maps (Grad-CAMs). Then, $F_G$ is transformed to the same shape of $F_{cls}$ and added with it to obtain $F_{DINO}$. $F_C$ is also transformed to $F_{Res}$, aligning its dimension to $F_{DINO}$. Next, $F_{Res}$ and $F_{DINO}$ are concatenated and sent to a gating network, which generates two learnable weight vectors $A_{Res}$ and $A_{DINO}$ for the element-wise fusion of $F_{Res}$ and $F_{DINO}$. The fused feature $F_{Fus}$ is finally fed into a classifier for diagnostic prediction.}
\label{sup:fig1}
\end{figure}

\begin{figure}[htbp]
\centering
\includegraphics[width=0.91\textwidth]{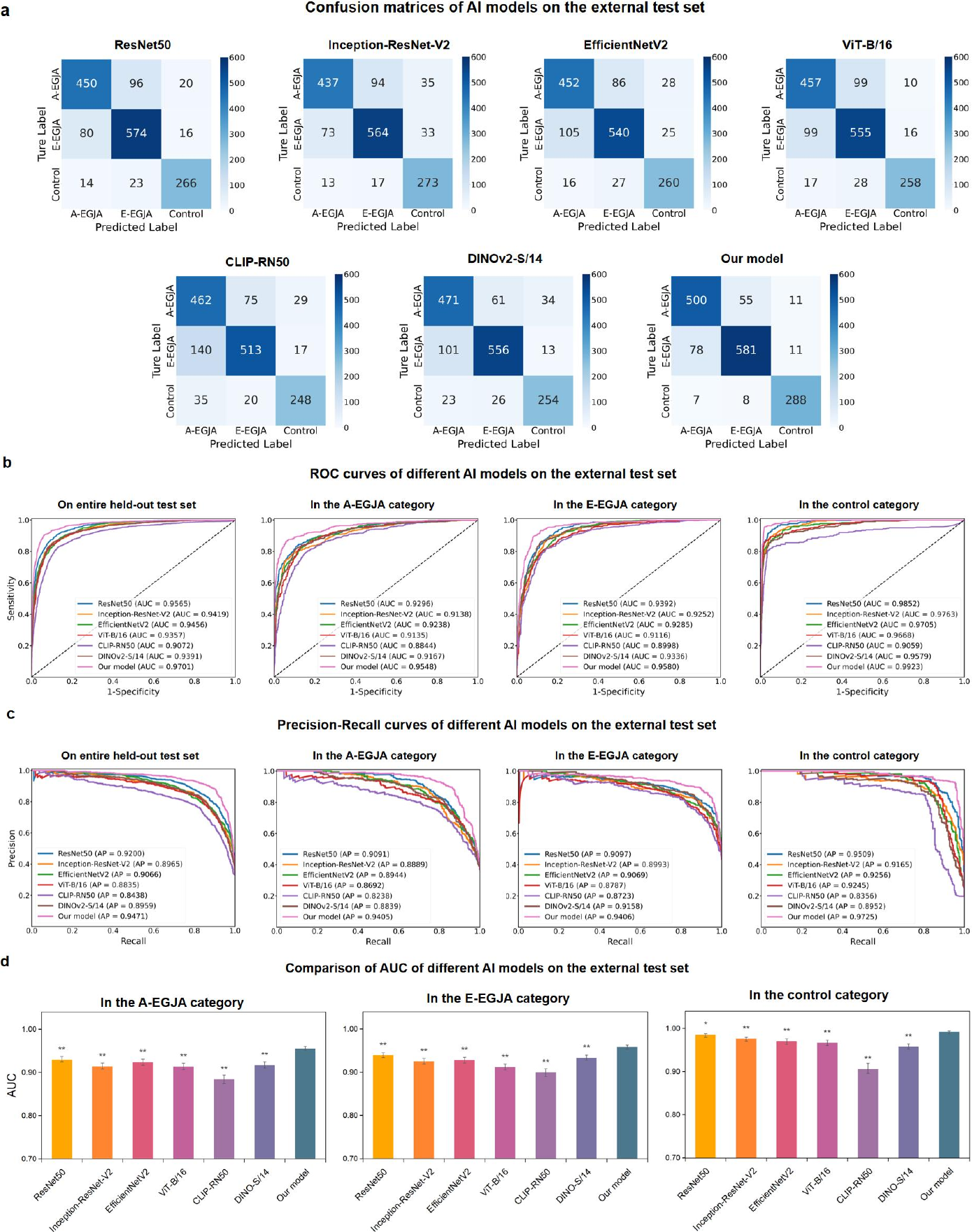}
\caption{Classification performance of different AI models on the external test set at the image level. \textbf{a:} the classification confusion matrices of different models. Numbers in each matrix represent the count of images in each category classified correctly (diagonal) and incorrectly (off the diagonal). \textbf{b:} the ROC curves of different models. The first column shows the ROC curves on the entire test set while the second to last columns represent the curves across categories (A-EGJA, E-EGJA, control) of the test set. \textbf{c:} the precision-recall curves of different models. The first column shows the precision-recall curves on the entire test set while the second to last columns represent the curves across categories of the test set, respectively. \textbf{d:} the comparison of different AI models' AUC across categories. * indicates P < 0.05, ** indicates P < 0.01, ns indicates P > 0.05 for comparison of AUC with our model using the Delong test.}
\label{sup:fig2}
\end{figure}

\begin{figure}[htbp]
\centering
\includegraphics[width=0.90\textwidth]{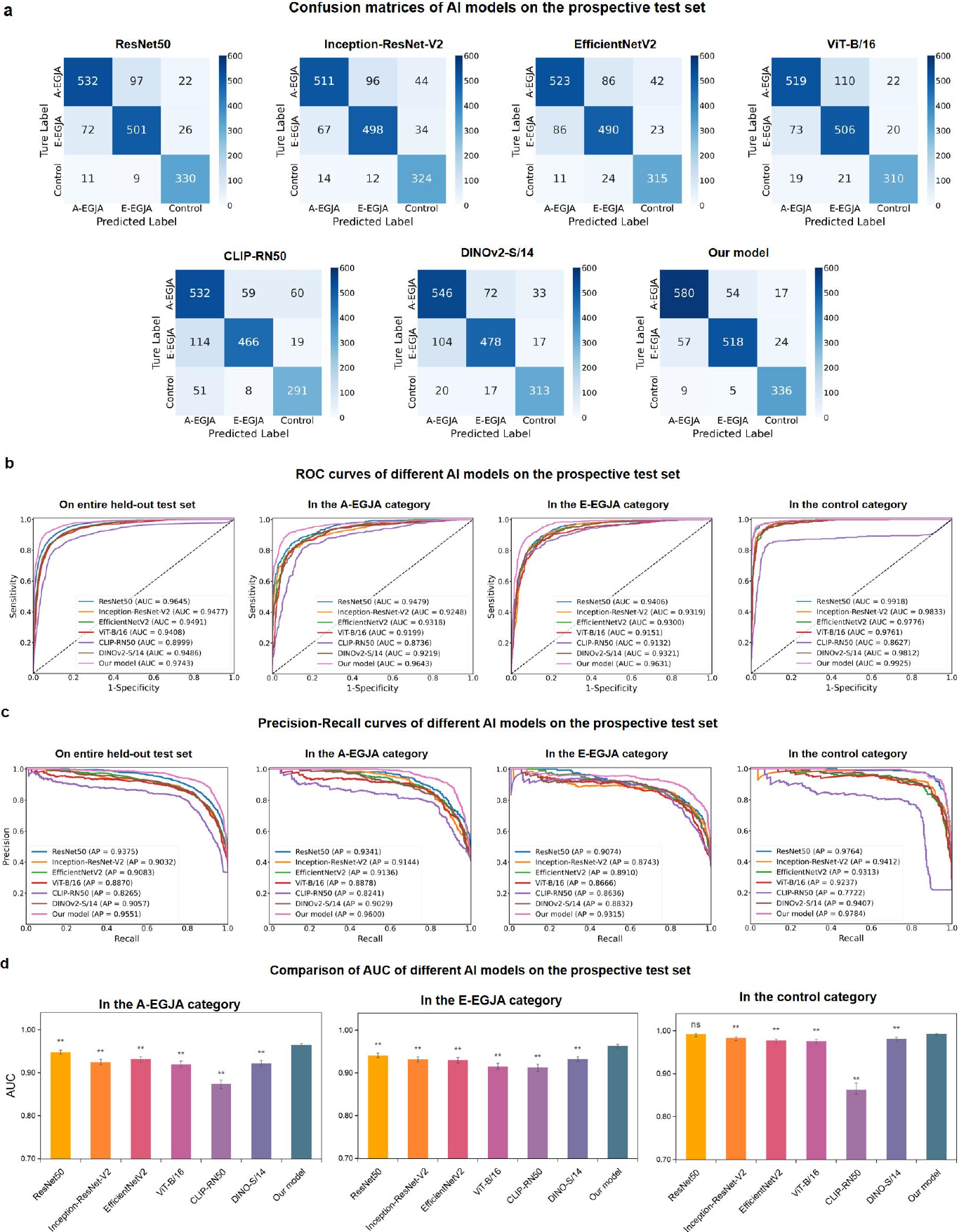}
\caption{Classification performance of different AI models on the prospective test set at the image level. \textbf{a:} the classification confusion matrices of different models. Numbers in each matrix represent the count of images in each category classified correctly (diagonal) and incorrectly (off the diagonal). \textbf{b:} the ROC curves of different models. The first column shows the ROC curves on the entire test set while the second to last columns represent the curves across categories (A-EGJA, E-EGJA, control) of the test set. \textbf{c:} the precision-recall curves of different models. The first column shows the precision-recall curves on the entire test set while the second to last columns represent the curves across categories of the test set, respectively. \textbf{d:} the comparison of different AI models' AUC across categories. * indicates P < 0.05, ** indicates P < 0.01, ns indicates P > 0.05 for comparison of AUC with our model using the Delong test.}
\label{sup:fig3}
\end{figure}

\begin{figure}[htbp]
\centering
\includegraphics[width=0.95\textwidth]{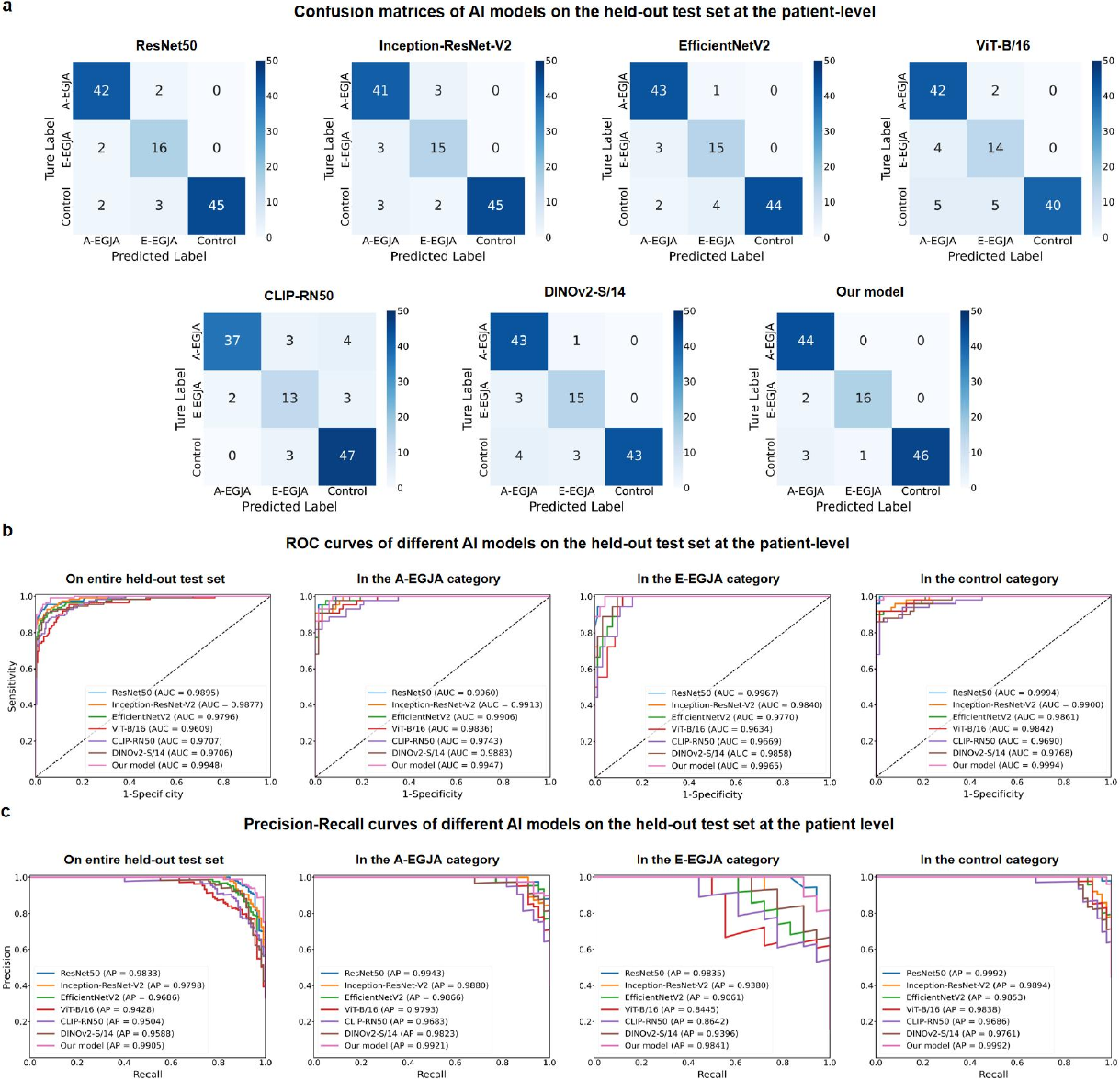}
\caption{Classification performance of different AI models on the held-out test set at the patient level. \textbf{a:} the classification confusion matrices of different models. Numbers in each matrix represent the count of images in each category classified correctly (diagonal) and incorrectly (off the diagonal). \textbf{b:} the ROC curves of different models. The first column shows the ROC curves on the entire test set while the second to last columns represent the curves across categories (A-EGJA, E-EGJA, control) of the test set. \textbf{c:} the precision-recall curves of different models. The first column shows the precision-recall curves on the entire test set while the second to last columns represent the curves across categories of the test set, respectively.}
\label{sup:fig4}
\end{figure}

\begin{figure}[htbp]
\centering
\includegraphics[width=0.7\textwidth]{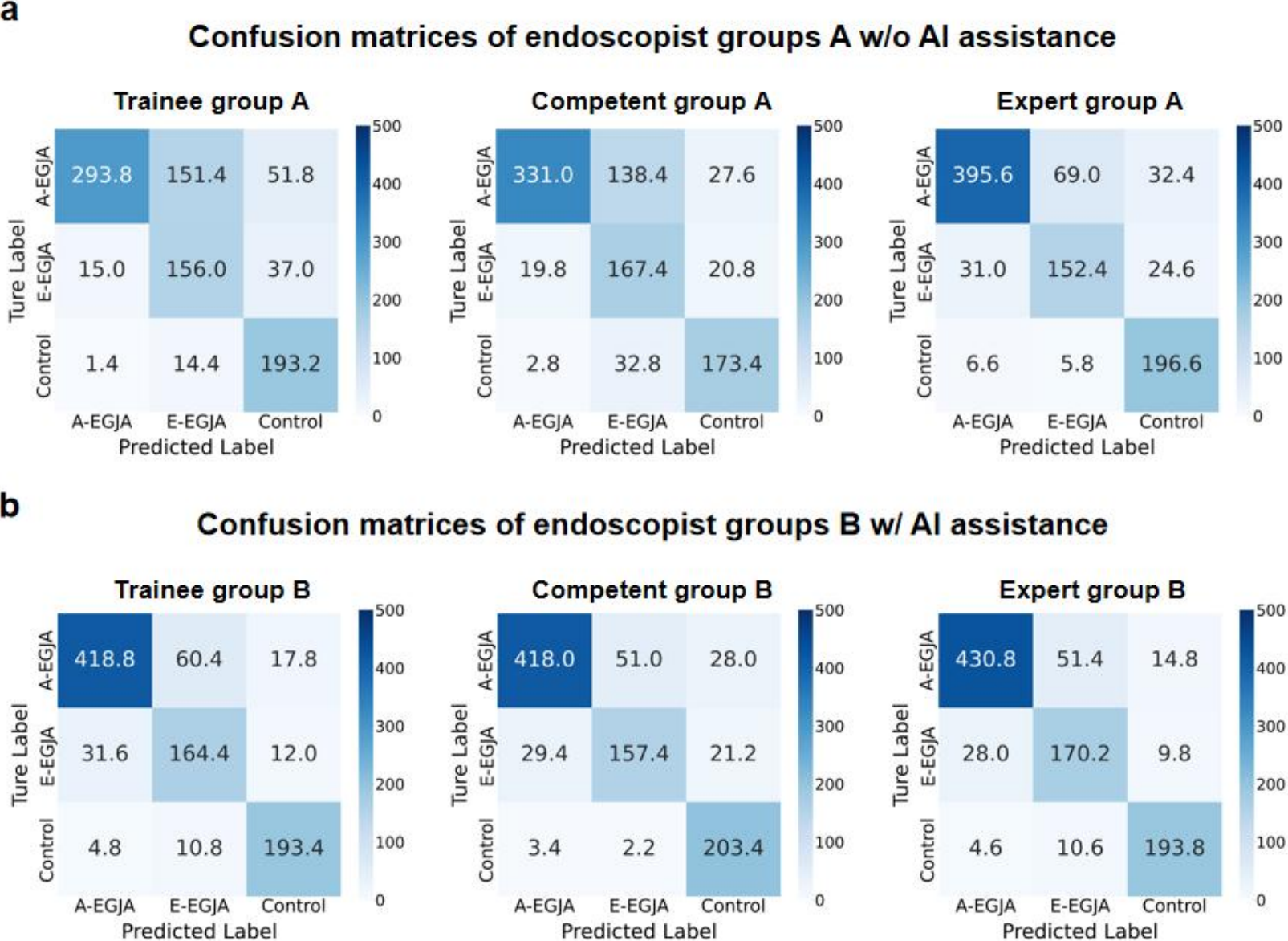}
\caption{Classification confusion matrices of different endoscopist groups on the held-out test set at the image level. \textbf{a:} the performance of each group A without the assistance of our model. \textbf{b:} the performance of each group B with the assistance of our model. Numbers in each matrix represent the count of images in each category classified correctly (diagonal) and incorrectly (off the diagonal).}
\label{sup:fig5}
\end{figure}

As mentioned in the Methods, our model consists of two encoders (ResNet50 \cite{ref26} and DINOv2 \cite{ref25}), a gating network, and a classifier (Fig.~\ref{sup:fig1}). Below we describe each as follows.

\textbf{ResNet50 encoder:} we employ ResNet50 \cite{ref26}, a well-established convolutional neural network (CNN), as a base module for local feature extraction. ResNet50 includes a large number of convolutional layers whose limited receptive fields make them more sensitive and proficient at extracting local details of images. The structure of ResNet50 mainly includes 16 basic residual modules, each has 3 convolutional layers with a residual connection, playing a major role in feature extraction. We remove the original fully-connected layer for classification from ResNet50 so that the output of our ResNet50 encoder is the feature map of the last residual module, denoted by $F_C\in R^{H_{Res}\times W_{Res}\times C_{Res} }$, where $H_{Res}$, $W_{Res}$, and $C_{Res}$ represent the height, width, and channel of $F_{C}$, respectively. It is worth noting that the parameters of ResNet50 are pre-trained on the ImageNet dataset \cite{ref35}.

\textbf{DINOv2 Encoder:} We employ DINOv2 \cite{ref25} as another encoder. DINOv2, as an improved version of DINO \cite{ref50}, is one of the cutting-edge vision foundation models. Specifically, its model structure is based on the vision transformer (ViT) \cite{ref38} that is powerful to capture global features. It has been pre-trained to acquire robust and sufficient knowledge from massive amounts of unlabeled images. The largest pre-trained DINOv2 is the ViT-G/14 with 1B parameters, however, considering its high computational cost, here we use its distilled model (ViT-S/14). It mainly includes a patch embedding layer, a position embedding layer, and 12 ViT blocks. The input image of this encoder is firstly embedded through a patch embedding layer and a position embedding layer, during which a class token is concatenated into the embeddings. Subsequently, the global visual features can be effectively extracted via the built-in multi-head self-attention mechanism in each ViT block, which are denoted by $F_G\in R^{H_{DINO}\times W_{DINO}\times C_{DINO}}$ and $F_cls\in R^{1\times C_{DINO}}$. $H_{DINO}$, $W_{DINO}$, and $C_{DINO}$ denote the height, width, and channel of $F_{G}$, respectively. $F_{cls}$ is the output class token, normally, it will be directly fed into a classifier to output the predicted results, but here we combine $F_{cls}$ with $F_{G}$, the output visual tokens, to improve the feature representation. Specifically, the reshape and average pooling operations are applied to align the dimension of  $F_{G}$ with $F_{cls}$. Then, the aligned  $F_{G}$ is added to $F_{cls}$ to get $F_{DINO}\in R^{1\times C_{DINO}}$ as an integrated feature from our DINOv2 encoder.

\textbf{Gating network in the mixture-of-experts (MoE):} After extracting features from the ResNet50 and DINOv2 encoders, we built a gating network to implement MoE \cite{ref27} for feature fusion. MoE was first proposed by Jacobs et al. \cite{ref51} to deal with the hidden heterogeneity in data. It has been widely used in the fields of natural language processing and computer vision and has achieved impressive performance \cite{ref52}. In general, MoE can harness multiple experts (represented as sub-networks in AI models) and exploit their collective knowledge. Its essential process is to dynamically assign weights to each expert through a gating network, so as to effectively fuse the knowledge of different experts to produce more comprehensive output. Here we apply the principle of MoE to fuse $F_{DINO}$ and $F_{Res}$ at the element level. As shown in Fig.~\ref{sup:fig1}, the gating network outputs two learnable weight vectors $A_{Res}\in R^{1\times C_{DINO}}$ and $A_{DINO}\in R^{1\times C_{DINO}}$ to element-wisely fuse $F_{DINO}$ and $F_{Res}$. Compared with original vector-level fusion in MoE, our design enables a fine fusion. It is worth noting that $F_{Res}$ is obtained from $F_{C}$ after passing through a pooling layer and a $1 \times 1$ convolutional layer to align its dimensions with $F_{DINO}$. As a result, both $F_{DINO}$ and $F_{Res}$ are of size $1 \times C_{DINO}$, and they are concatenated to a $1 \times C_{DINO} \times 2$ sized input to the gating network. There are three one-dimensional convolutional layers in the gating network, and the first two layers are followed by a ReLU layer and a dropout layer. Finalised with a softmax layer, the gating network generates a $1 \times C_{DINO} \times 2$ sized output, which is further split into two weight vectors $A_{DINO}$ and $A_{Res}$ of size $1 \times C_{DINO}$. The sum of the corresponding elements in the two vectors is 1. They serve as weighting coefficients to add $F_{DINO}$ and $F_{Res}$ into the fused feature $F_{Fus} \in R^{1\times C_{DINO}}$.

\textbf{Classifier:} After feature fusion, $F_{Fus}$ is fed into a classifier that mainly includes a fully-connected layer. It generates the diagnostic predictions for each category (E-EGJA, A-EGJA, and control), and finally outputs the model diagnosis for the input image.

\section*{Data standardisation}

Our training data were collected from 6 hospitals. To address the centre-specific variability, several essential data standardisation steps, including image normalisation and image augmentation (random cropping, rotation, and colour jittering), are conducted. In the Method Section of our paper, we have offered our default model (image-level training, denoted by Ours) and its two variants (patient-level training, denoted by Ours-v1 and Ours-v2). To justify the data standardisation method, we additionally present three baseline models, Ours-Base, Ours-v1-Base, and Ours-v2-Base, which were trained without data standardisation. The results in Table~\ref{sup:tab13} and \ref{sup:tab14} show both image-level and patient-level evaluations across each hospital. It can be seen that Ours, Ours-v1, and Ours-v2 consistently outperform their respective baselines, attesting the effectiveness of the standardisation process.

\section*{Inclusion of NBI and WLI images}

Our model is developed based on the deep learning technique, which implicitly learns the discriminative and power visual features of lesions through the optimisation against their labels. Therefore, it is sufficient for our model to only use WLI images for EGJA diagnosis. Nevertheless, the NBI can further enhance the microvascular and microstructural contrast of the mucosal surface, making the lesion area more visible to be diagnosed. Therefore, we include both WLI and NBI images for our model training, enhancing its feature extraction ability and generalisability of our model. To validate it, we separate the NBI images and WLI images from our training set (8,249 images) and held-out test set (914 images), and we construct the WLI training set that comprises 7,211 WLI images from the training set and the WLI test set that comprises 810 WLI images from the held-out test set. Then, we use the WLI training set to train the WLI version of our model (denoted by Ours-WLI) with the same training configuration as our original model (denoted by Ours), and we test the performance on the WLI test set. The results in Table~\ref{sup:tab15} below show that our model (Ours-WLI) can achieve already satisfactory performance (accuracy: 0.9160) when training with only WLI images and can be further enhanced when adding additional NBI images (accuracy: 0.9333).

\clearpage
\section*{Supplementary figures}

\begin{figure}[htbp]
\centering
\includegraphics[width=0.95\textwidth]{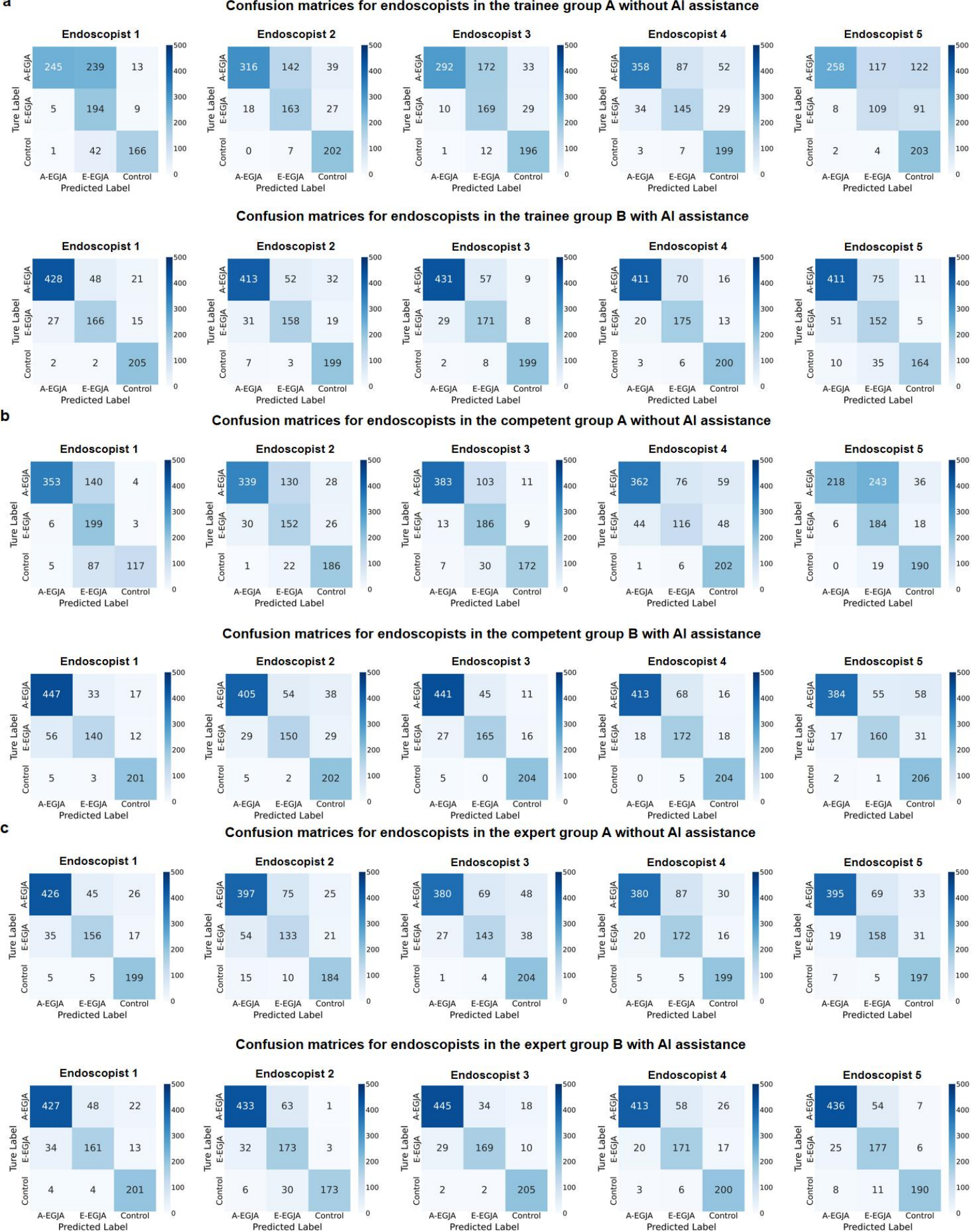}
\caption{Classification confusion matrices of endoscopists on the held-out test set at the image level. \textbf{a:} confusion matrices for endoscopists in the trainee groups A and B. \textbf{b:} confusion matrices for endoscopists in the competent groups A and B. \textbf{c:} confusion matrices for endoscopists in the expert groups A and B. Numbers in matrices represent the count of images in each category classified correctly (diagonal) and incorrectly (off the diagonal).}
\label{sup:fig6}
\end{figure}

\begin{figure}[htbp]
\centering
\includegraphics[width=0.95\textwidth]{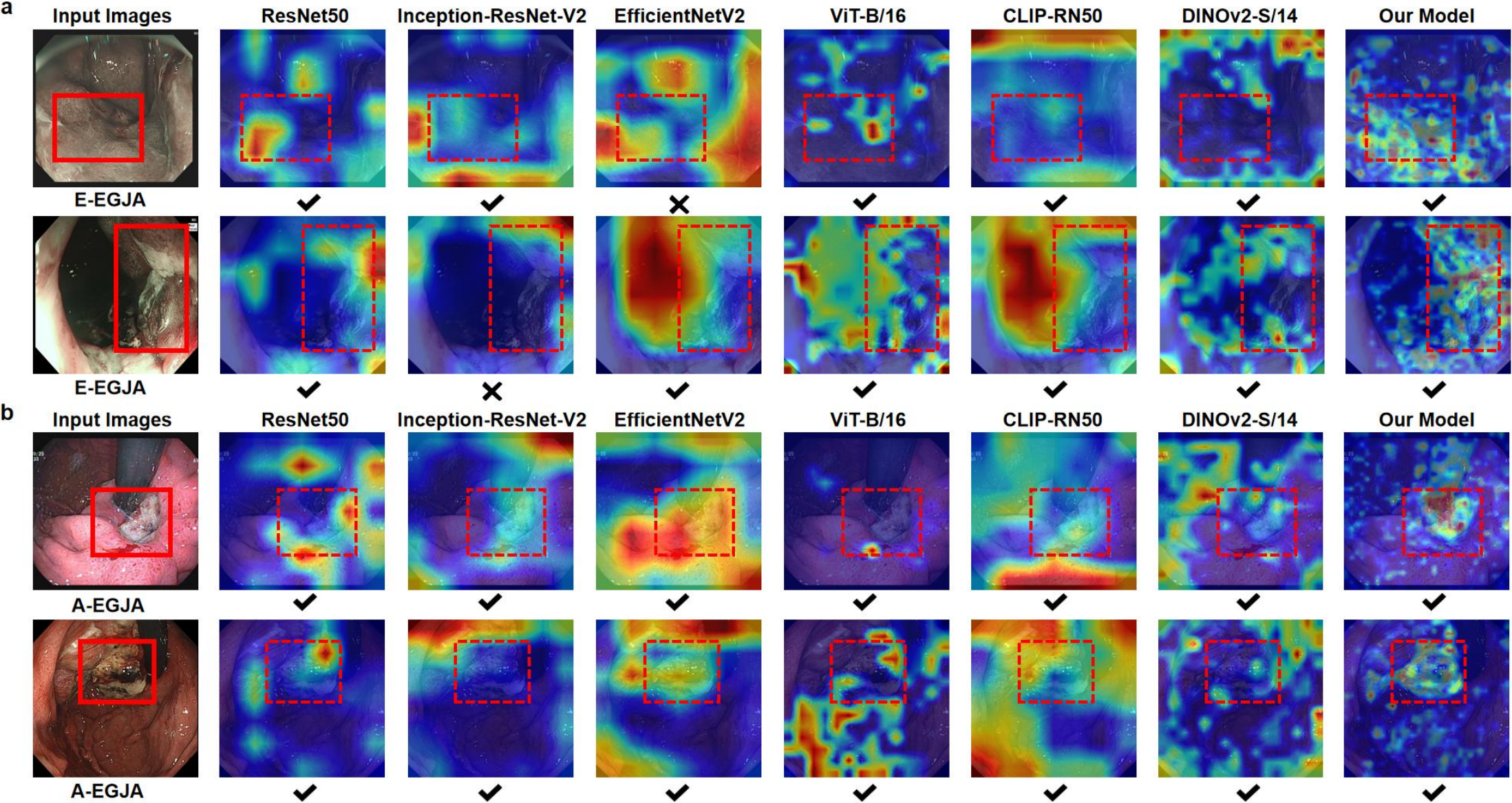}
\caption{Comparison of Grad-CAMs by each AI model in diagnosing the E-EGJA \textbf{(a)} and A-EGJA \textbf{(b)} on the external test set. The colours from blue to red in each Grad-CAM denote the activation values from low to high, depending on which the model makes the prediction. The red bounding boxes in input images are annotated by chief endoscopists.}
\label{sup:fig7}
\end{figure}

\begin{figure}[htbp]
\centering
\includegraphics[width=0.95\textwidth]{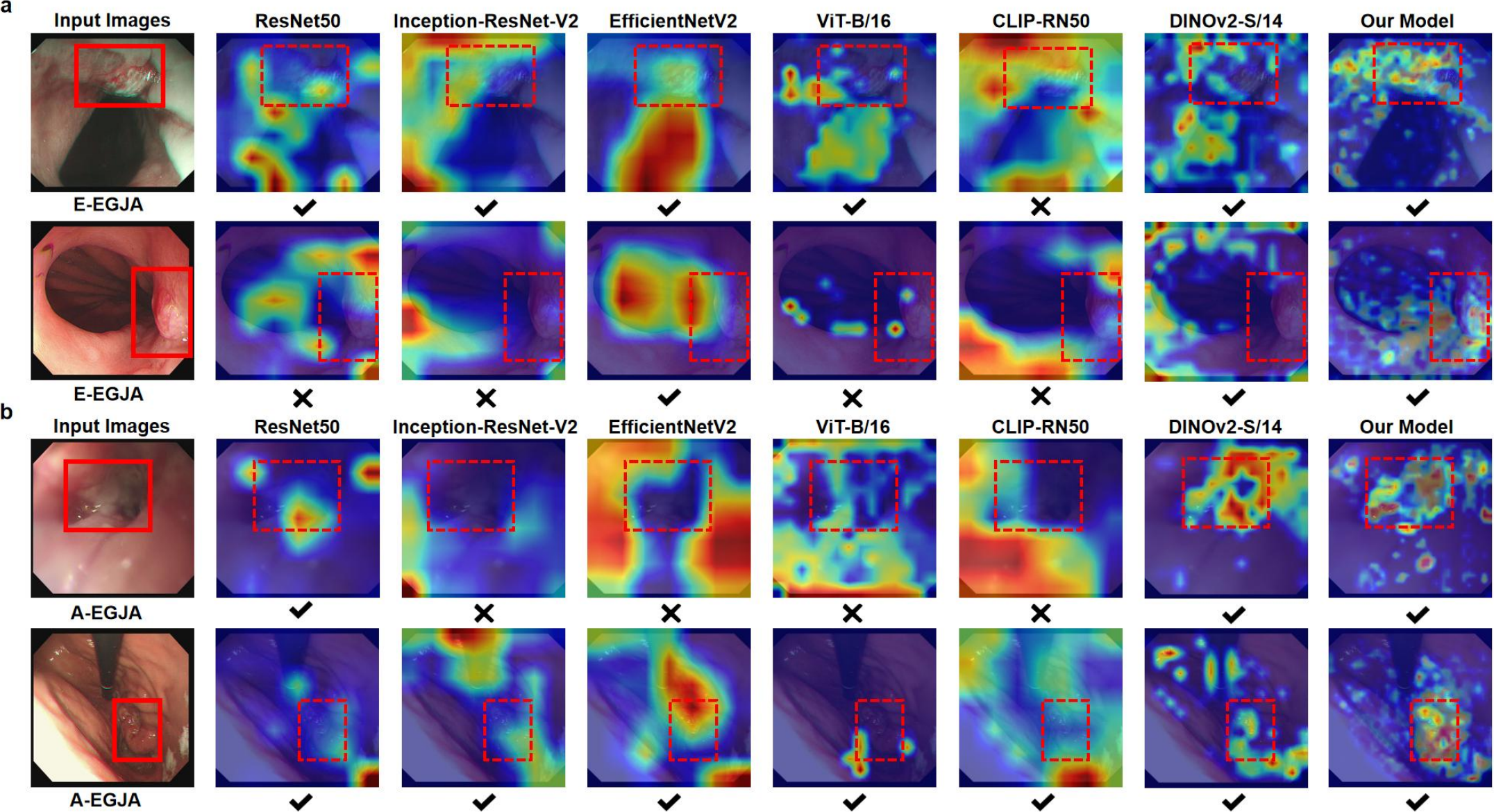}
\caption{Comparison of Grad-CAMs by each AI model in diagnosing the E-EGJA \textbf{(a)} and A-EGJA \textbf{(b)} on the prospective test set. The colours from blue to red in each Grad-CAM denote the activation values from low to high, depending on which the model makes the prediction. The red bounding boxes in input images are annotated by chief endoscopists.}
\label{sup:fig8}
\end{figure}

\begin{figure}[htbp]
\centering
\includegraphics[width=0.95\textwidth]{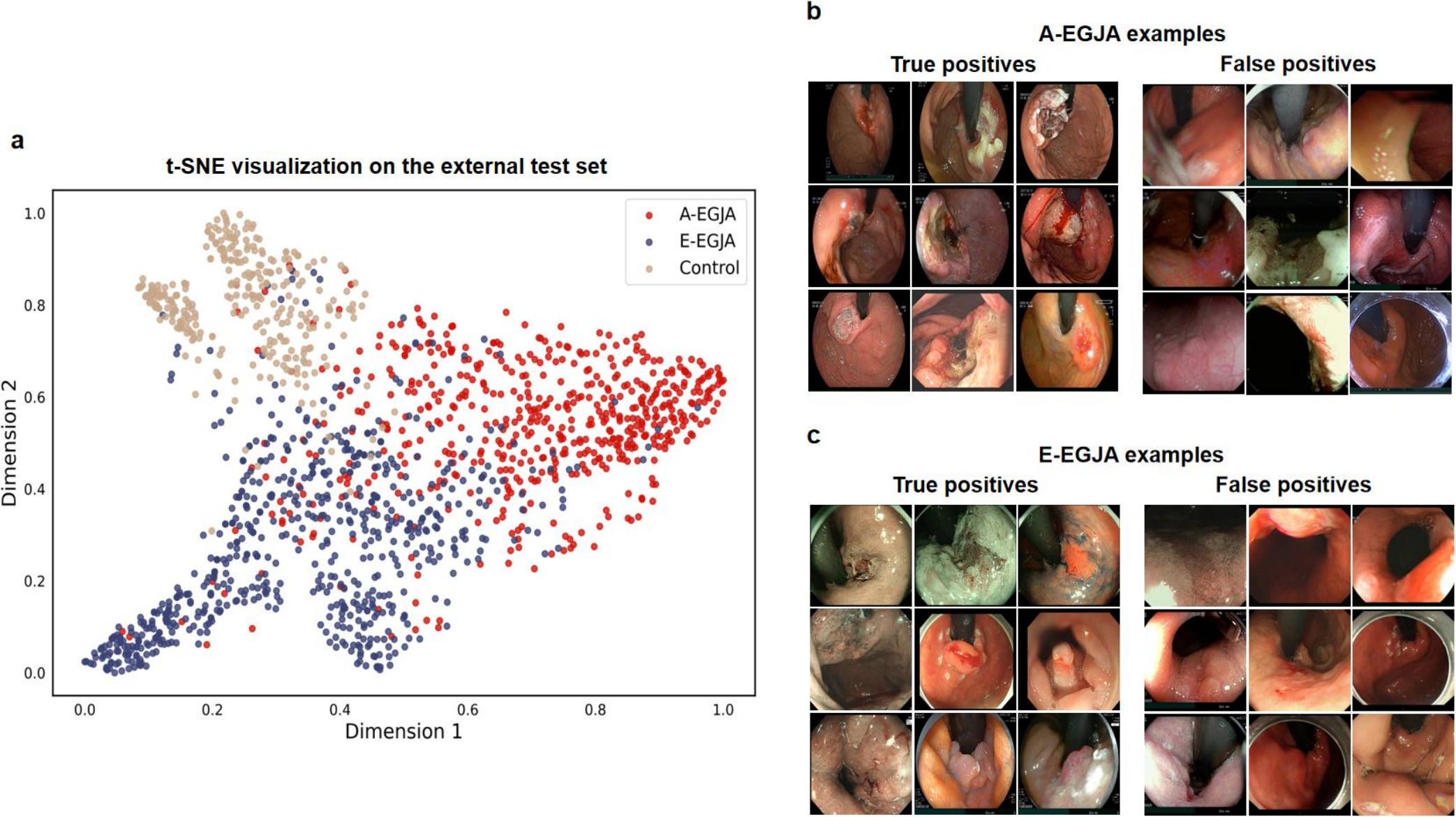}
\caption{The t-SNE visualisation of our model (left) and image examples diagnosed by our model as EGJA on the external test set. \textbf{a:} the two-dimensional visualisation of the fused feature $F_{Fus}$ of our model using t-SNE. \textbf{b:} A-EGJA examples diagnosed by our model. \textbf{c:} E-EGJA examples diagnosed by our model.}
\label{sup:fig9}
\end{figure}

\begin{figure}[htbp]
\centering
\includegraphics[width=0.95\textwidth]{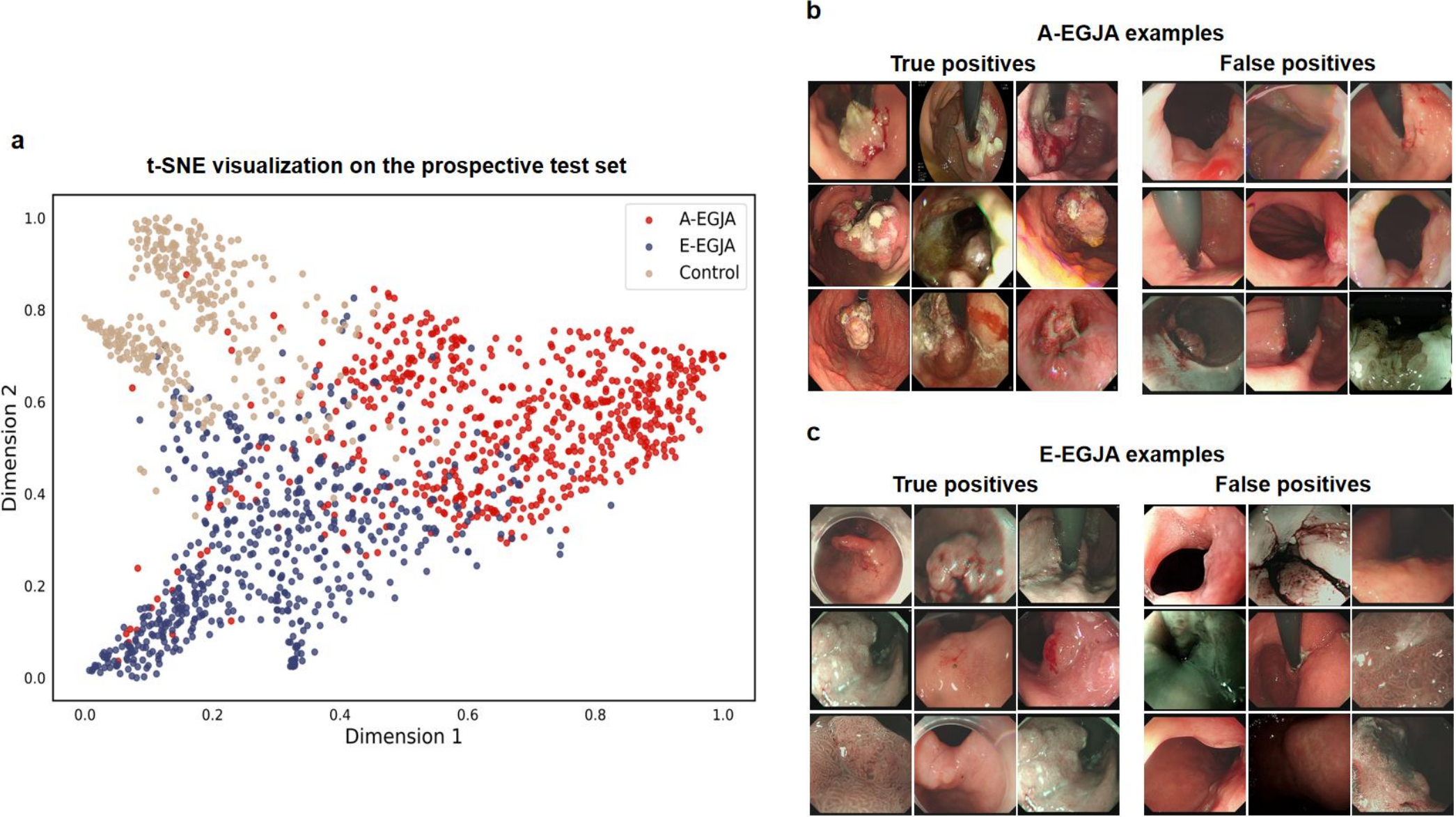}
\caption{The t-SNE visualisation of our model (left) and image examples diagnosed by our model as EGJA on the prospective test set. \textbf{a:} the two-dimensional visualisation of the fused feature $F_{Fus}$ of our model using t-SNE. \textbf{b:} A-EGJA examples diagnosed by our model. \textbf{c:} E-EGJA examples diagnosed by our model.}
\label{sup:fig10}
\end{figure}

\clearpage
\section*{Supplementary tables}

\vspace{180pt}

\begin{table}[htbp]
\centering
\scriptsize
\caption{Overall comparison to representative AI models on the training set at the image level using five-fold cross validation.}
\vspace{5pt}
\begin{tabular}{cccccccc}
\toprule
& \textbf{ResNet50} & \textbf{Inception-ResNet-V2} & \textbf{EfficientNetV2} & \textbf{ViT-B/16} & \textbf{CLIP-RN50} & \multicolumn{1}{c}{\textbf{DINOv2-S/14}} & \textbf{Our model} \\
\midrule
\multicolumn{1}{c}{Accuracy} & 0.9257 & 0.9149 & 0.9092 & 0.9111 & 0.8833 & 0.9144 & \textbf{0.9423} \\
\multicolumn{1}{c}{(95\% CI)} & (0.9200-0.9313) & (0.9088-0.9209) & (0.9030-0.9154) & (0.9050-0.9172) & (0.8764-0.8903) & \multicolumn{1}{c}{(0.9083-0.9204)} & (0.9372-0.9473) \\ \addlinespace[0.4em]
\multicolumn{1}{c}{Sensitivity} & 0.9274 & 0.9049 & 0.9017 & 0.9011 & 0.8901 & 0.9066 & \textbf{0.9398} \\
\multicolumn{1}{c}{(95\% CI)} & \multicolumn{1}{c}{(0.9177-0.9371)} & \multicolumn{1}{c}{(0.8940-0.91580)} & (0.8906-0.9127) & (0.8899-0.9122) & (0.8786-0.9015) & \multicolumn{1}{c}{(0.8959-0.9172)} & (0.9310-0.9487) \\ \addlinespace[0.4em]
\multicolumn{1}{c}{Specificity} & 0.9595 & 0.9490 & 0.9471 & 0.9472 & 0.9382 & 0.9504 & \textbf{0.9671} \\
\multicolumn{1}{c}{(95\% CI)} & (0.9542-0.9647) & (0.9433-0.9547) & (0.9413-0.9530) & (0.9414-0.9530) & (0.9317-0.9447) & \multicolumn{1}{c}{(0.9446-0.9561)} & (0.9623-0.9718) \\ \addlinespace[0.4em]
\multicolumn{1}{c}{PPV} & 0.9179 & 0.9151 & 0.9055 & 0.9101 & 0.8689 & 0.9102 & \textbf{0.9389} \\
\bottomrule
\end{tabular}%
\label{sup:tab1}
\end{table}

\begin{table}[h]
\centering
\scriptsize
\caption{Comparison to representative AI models across categories (A-EGJA, E-EGJA, and control) on the training set using five-fold cross validation.}
\vspace{5pt}
    \begin{tabular}{cccccccc}
    \toprule
     & \textbf{ResNet50} & \textbf{Inception-ResNet-V2} & \textbf{EfficientNetV2} & \textbf{ViT-B/16} & \textbf{CLIP-RN50} & {\textbf{DINOv2-S/14}} & \textbf{Our model} \\
    \midrule
    \multicolumn{8}{c}{\textbf{Classification results in the A-EGJA category}} \\
    {Accuracy} & 0.9305 & 0.9193 & 0.9155 & 0.9170 & 0.8921 & 0.9220 & \textbf{0.9457} \\
    {(95\% CI)} & (0.9250-0.9360) & (0.9135-0.9252) & (0.9095-0.9215) & (0.9111-0.9230) & (0.8854-0.8988) & {(0.9162-0.9278)} & \textbf{(0.9408-0.9506)} \\ \addlinespace[0.4em]
    {Sensitivity} & 0.9237 & 0.9401 & 0.9288 & 0.9365 & 0.8712 & 0.9349 & \textbf{0.9495} \\
    {(95\% CI)} & {(0.9159-0.9315)} & {(0.9331-0.9470)} & (0.9213-0.9364) & (0.9293-0.9436) & (0.8613-0.8810) & {(0.9277-0.9421)} & \textbf{(0.9431-0.9559)} \\ \addlinespace[0.4em]
    {Specificity} & 0.9385 & 0.8950 & 0.8997 & 0.8942 & 0.9166 & 0.9069 & \textbf{0.9412} \\
    {(95\% CI)} & (0.9309-0.9462) & (0.8852-0.9047) & (0.8902-0.9093) & (0.8844-0.9040) & (0.9078-0.9254) & {(0.8976-0.9161)} & \textbf{(0.9337-0.9487)} \\
\bottomrule
    \end{tabular}%
\label{sup:tab2}
\end{table}

\begin{table}[htbp]
\centering
\scriptsize
\caption{Comparison to representative AI models across categories (A-EGJA, E-EGJA, and control) on the held-out test set at the image level.}
\vspace{5pt}
    \begin{tabular}{cccccccc}
    \toprule
    & \textbf{ResNet50} & \textbf{Inception-ResNet-V2} & \textbf{EfficientNetV2} & \textbf{ViT-B/16} & \textbf{CLIP-RN50} & {\textbf{DINOv2-S/14}} & \textbf{Our model} \\
    \midrule
    \multicolumn{8}{c}{\textbf{Classification results in the A-EGJA category}} \\
    {Accuracy} & 0.9179 & 0.8939 & 0.8840 & 0.8862 & 0.8687 & 0.8939 & \textbf{0.93} \\
    {(95\% CI)} & (0.9002-0.9357) & (0.8739-0.9138) & (0.8633-0.9048) & (0.8656-0.9068) & (0.8468-0.8906) & {(0.8739-0.9138)} & (0.9134-0.9465) \\ \addlinespace[0.4em]
    {Sensitivity} & 0.9115 & 0.8672 & 0.9115 & 0.9256 & 0.8551 & 0.9336 & \textbf{0.9316} \\
    {(95\% CI)} & {(0.8865-0.9364)} & {(0.8374-0.8970)} & (0.8865-0.9364) & (0.9025-0.9486) & (0.8242-0.8861) & {(0.9117-0.9555)} & (0.9094-0.9538) \\ \addlinespace[0.4em]
    {Specificity} & 0.9257 & 0.9257 & 0.8513 & 0.8393 & 0.8849 & 0.8465 & \textbf{0.9281} \\
    {(95\% CI)} & (0.9005-0.9508) & (0.9005-0.9508) & (0.8172-0.8855) & (0.8041-0.8746) & (0.8543-0.9155) & {(0.8119-0.8811)} & (0.9033-0.9529) \\
\bottomrule
    \end{tabular}%
\label{sup:tab3}
\end{table}

\begin{table}[htbp]
\centering
\scriptsize
\caption{Comparison to representative AI models across categories on the external test set at the image level.}
\vspace{5pt}
    \begin{tabular}{cccccccc}
    \toprule
          & \textbf{ResNet50} & \textbf{Inception-ResNet-V2} & \textbf{EfficientNetV2} & \textbf{ViT-B/16} & \textbf{CLIP-RN50} & {\textbf{DINOv2-S/14}} & \textbf{Our model} \\
    \midrule
    \multicolumn{8}{c}{\textbf{Classification results in the A-EGJA category}} \\
    {Accuracy} & 0.8635 & 0.8603 & 0.8473 & 0.8538 & 0.8187 & 0.8577 & \textbf{0.9019} \\
    {(95\% CI)} & (0.8464-0.8807) & (0.843-0.8776) & (0.8293-0.8653) & (0.8361-0.8715) & (0.7995-0.838) & {(0.8402-0.8752)} & (0.887-0.9167) \\ \addlinespace[0.4em]
    {Sensitivity} & 0.7951 & 0.7721 & 0.7986 & 0.8074 & 0.8163 & 0.8322 & \textbf{0.8834} \\
    {(95\% CI)} & {(0.7618-0.8283)} & {(0.7375-0.8066)} & (0.7655-0.8316) & (0.7749-0.8399) & (0.7843-0.8482) & {(0.8014-0.8629)} & (0.857-0.9098) \\ \addlinespace[0.4em]
    {Specificity} & 0.9034 & 0.9116 & 0.8756 & 0.8808 & 0.8201 & 0.8726 & \textbf{0.9126} \\
    {(95\% CI)} & (0.8848-0.922) & (0.8938-0.9294) & (0.8549-0.8964) & (0.8604-0.9011) & (0.796-0.8443) & {(0.8516-0.8935)} & (0.8949-0.9304) \\
\bottomrule
    \end{tabular}%
\label{sup:tab4}
\end{table}

\begin{table}[htbp]
\centering
\scriptsize
\caption{Comparison to representative AI models across categories on the prospective test set at the image level.}
\vspace{5pt}
    \begin{tabular}{cccccccc}
    \toprule
          & \textbf{ResNet50} & \textbf{Inception-ResNet-V2} & \textbf{EfficientNetV2} & \textbf{ViT-B/16} & \textbf{CLIP-RN50} & {\textbf{DINOv2-S/14}} & \textbf{Our model} \\
    \midrule
    \multicolumn{8}{c}{\textbf{Classification results in the A-EGJA category}} \\
    {Accuracy} & 0.8738 & 0.8619 & 0.8594 & 0.8600  & 0.8225 & 0.8569 & \textbf{0.9144} \\
    {(95\% CI)} & (0.8575-0.89) & (0.845-0.8788) & (0.8423-0.8764) & (0.843-0.877) & (0.8038-0.8412) & {(0.8397-0.874)} & (0.9007-0.9281) \\ \addlinespace[0.4em]
    {Sensitivity} & 0.8172 & 0.7849 & 0.8034 & 0.7972 & 0.8172 & 0.8387 & \textbf{0.8909} \\
    {(95\% CI)} & {(0.7875-0.8469)} & {(0.7534-0.8165)} & (0.7728-0.8339) & (0.7663-0.8281) & (0.7875-0.8469) & {(0.8105-0.867)} & (0.867-0.9149) \\ \addlinespace[0.4em]
    {Specificity} & 0.9125 & 0.9146 & 0.8978 & 0.9031 & 0.8261 & 0.8693 & \textbf{0.9305} \\
    {(95\% CI)} & (0.8946-0.9305) & (0.8969-0.9324) & (0.8785-0.9171) & (0.8842-0.9219) & (0.802-0.8502) & {(0.8479-0.8908)} & (0.9143-0.9466) \\
\bottomrule
    \end{tabular}%
\label{sup:tab5}
\end{table}

\begin{table}[t]
\centering
\scriptsize
\caption{P values of different AI models' AUC compared with our model using the Delong test across categories on three test sets at the image level.}
\vspace{5pt}
    \begin{tabular}{ccccccccc}
    \toprule
    \textbf{Category} &       & \textbf{ResNet50} & \textbf{Inception-ResNet-V2} & \textbf{EfficientNetV2} & \textbf{ViT-B/16} & \textbf{CLIP-RN50} & \textbf{DINOv2-S/14} & \textbf{Our model} \\
    \midrule
    \multicolumn{9}{c}{\textbf{Held-out test set (n=914)}} \\
    \multicolumn{1}{c}{\multirow{2}[0]{*}{A-EGJA}} & AUC   & 0.9677 & 0.9585 & 0.9502 & 0.9527 & 0.9183 & \multicolumn{1}{c}{0.9607} & \textbf{0.9735} \\
          & P value & 0.04  & <0.0001 & <0.0001 & <0.0001 & <0.0001 & \multicolumn{1}{c}{0.0046} & Reference \\ \addlinespace[0.4em]
    \multicolumn{1}{c}{\multirow{2}[0]{*}{E-EGJA}} & {AUC} & 0.9461 & 0.9463 & 0.9418 & 0.9496 & 0.9174 & \multicolumn{1}{c}{0.9504} & \textbf{0.9547} \\
          & {P value} & 0.019 & 0.047 & 0.02  & 0.45  & 0.0004 & \multicolumn{1}{c}{0.49} & Reference \\ \addlinespace[0.4em] 
    \multicolumn{1}{c}{\multirow{2}[0]{*}{Control}} & AUC   & 0.9979 & 0.9979 & 0.9979 & 0.995 & 0.9951 & \multicolumn{1}{c}{0.9967} & \textbf{0.9982} \\
          & P value & 0.75  & 0.75  & 0.68  & 0.055 & 0.13  & \multicolumn{1}{c}{0.27} & Reference \\
\bottomrule
    \end{tabular}%
\label{sup:tab6}
\end{table}

% \vspace{-10em}

\begin{table}[p]
\setlength{\abovecaptionskip}{-120pt}
\centering
\scriptsize
\caption{Comparison of the original and weighted results of our model on the held-out test set at the image level.}
\vspace{5pt}
    \begin{tabular}{ccccccccc}
    \toprule
    \multirow{2}[0]{*}{} & \textbf{Accuracy} & \textbf{Sensitivity} & \textbf{Specificity} & \textbf{PPV} & \textbf{NPV} & \multicolumn{1}{c}{\multirow{2}[0]{*}{\textbf{Kappa}}} & \multirow{2}[0]{*}{\textbf{AUC}} & \multicolumn{1}{c}{\multirow{2}[0]{*}{\textbf{AP}}} \\
          & (95\% CI) & (95\% CI) & (95\% CI) & (95\% CI) & (95\% CI) &       &       &  \\
    \midrule
    \multirow{2}[1]{*}{\textbf{Original}} & 0.9256 & 0.9227 & 0.9581 & 0.9189 & 0.9570 & \multirow{2}[1]{*}{0.8763} & \multirow{2}[1]{*}{0.9818} & \multirow{2}[1]{*}{0.9675} \\
          & (0.9086-0.9426) & (0.8946-0.9509) & (0.9423-0.9738) & (0.8877-0.9501) & (0.9419-0.9721) &       &       &  \\ \addlinespace[0.4em]
    \multirow{2}[1]{*}{\textbf{Weighted}} & 0.9353 & 0.9152 & 0.9483 & 0.9189 & 0.9453 & \multirow{2}[1]{*}{0.8763} & \multirow{2}[1]{*}{0.9736} & \multirow{2}[1]{*}{0.9676} \\
          & (0.8898-0.9809) & (0.8636-0.9668) & (0.9074-0.9893) & (0.8695-0.9701) & (0.9032-0.9874) &       &       &  \\
    \bottomrule
    \end{tabular}%
\label{sup:tab7}
\end{table}

% \clearpage
% \vspace{}

\begin{table}[t]
\centering
\scriptsize
\caption{Comparison of the original and weighted results of our model on the held-out test set at the image level.}
\vspace{5pt}
    \begin{tabular}{cccccccc}
    \toprule
          & \textbf{ResNet50} & \textbf{Inception-ResNet-V2} & \textbf{EfficientNetV2} & \textbf{ViT-B/16} & \textbf{CLIP-RN50} & {\textbf{DINOv2-S/14}} & \textbf{Our model} \\
    \midrule
    \multicolumn{8}{c}{\textbf{Classification results in the A-EGJA category}} \\
    {Accuracy} & 0.9464 & 0.9196 & 0.9464 & 0.9018 & 0.9196 & 0.9286 & 0.9554 \\
    {(95\% CI)} & (0.9047-0.9881) & (0.8693-0.9700) & (0.9047-0.9881) & (0.8467-0.9569) & (0.8693-0.9700) & {(0.8809-0.9763)} & (0.9171-0.9936) \\ \addlinespace[0.4em]
    {Sensitivity} & 0.9545 & 0.9318 & 0.9773 & 0.9545 & 0.8409 & 0.9773 & 1 \\
    {(95\% CI)} & {(0.8930-1.000)} & {(0.8573-1.000)} & (0.9332-1.000) & (0.8930-1.000) & (0.7328-0.9490) & {(0.9332-1.000)} & (1.000-1.000) \\ \addlinespace[0.4em]
    {Specificity} & 0.9412 & 0.9118 & 0.9265 & 0.8676 & 0.9706 & 0.8971 & 0.9265 \\
    {(95\% CI)} & (0.8853-0.9971) & (0.8443-0.9792) & (0.8644-0.9885) & (0.7871-0.9482) & (0.9304-1.000) & {(0.8248-0.9693)} & (0.8644-0.9885) \\
\bottomrule
    \end{tabular}%
\label{sup:tab8}
\end{table}

\begin{table}[htbp]
\centering
\scriptsize
\caption{Performance of our model in different patient groups (sex or age) on the held-out test set at the patient level.}
\vspace{5pt}
    \begin{tabular}{ccccccccc}
    \toprule
    \multirow{2}[0]{*}{} & \textbf{Accuracy} & \textbf{Sensitivity} & \textbf{Specificity} & \textbf{PPV} & \textbf{NPV} & \multicolumn{1}{c}{\multirow{2}[0]{*}{\textbf{Kappa}}} & \multirow{2}[0]{*}{\textbf{AUC}} & \multicolumn{1}{c}{\multirow{2}[0]{*}{\textbf{AP}}} \\
          & (95\% CI) & (95\% CI) & (95\% CI) & (95\% CI) & (95\% CI) &       &       &  \\
    \midrule
    \multicolumn{9}{c}{\textbf{Patient groups of different sex (male or female)}} \\
    \textbf{Male} & 0.9595 & 0.9444 & 0.9753 & 0.9615 & 0.9824 & \multirow{2}[0]{*}{0.9326} & \multirow{2}[0]{*}{0.9957} & \multirow{2}[0]{*}{0.9925} \\
    \textbf{(patients = 74)} & (0.9145-1.000) & (0.8585-1.000) & (0.9387-1.000) & (0.8975-1.000) & (0.9553-1.000) &       &       &  \\ \addlinespace[0.4em]
    \textbf{Female} & 0.9211 & 0.8667 & 0.9697 & 0.8750 & 0.9241 & \multirow{2}[0]{*}{0.7957} & \multirow{2}[0]{*}{0.9952} & \multirow{2}[0]{*}{0.9903} \\
    \textbf{(patients = 38)} & (0.8353-1.000) & (0.6591-1.000) & (0.937-1.000) & (0.7632-0.9868) & (0.8235-1.000) &       &       &  \\ \midrule
    \multicolumn{9}{c}{\textbf{Patient groups of different ages (age < 60, 60 $\leq$ age < 70, and age $\geq$ 70)}} \\
\bottomrule
    \end{tabular}%
\label{sup:tab9}
\end{table}

\begin{table}[htbp]
\centering
\scriptsize
\caption{Classification results of each endoscopist in the trainee groups A and B on the held-out test set at the image level.}
\vspace{5pt}
    \begin{tabular}{cccccc}
    \toprule
    \multicolumn{6}{c}{\textbf{Trainee group A w/o AI assistance}} \\
    \multicolumn{1}{c}{} & \textbf{Endoscopist 1} & \textbf{Endoscopist 2} & \textbf{Endoscopist 3} & \textbf{Endoscopist 4} & \textbf{Endoscopist 5} \\
    \midrule
    Accuracy & 0.6619 & 0.7188 & 0.7681 & 0.7451 & 0.6236 \\
    (95\% CI) & {(0.6313-0.6926)} & {(0.6897-0.7480)} & (0.7407-0.7954) & (0.7168-0.7733) & (0.5922-0.6550) \\ \addlinespace[0.4em]
    Sensitivity & 0.7400  & 0.7793 & 0.7899 & 0.7953 & 0.6715 \\
    (95\% CI) & (0.6957-0.7842) & (0.7363-0.8223) & (0.7462-0.8335) & (0.7544-0.8362) & (0.6267-0.7163) \\ \addlinespace[0.4em]
    Specificity & 0.8521 & 0.8750 & 0.8877 & 0.8841 & 0.8342 \\
    (95\% CI) & (0.8320-0.8723) & (0.8521-0.8979) & (0.8624-0.9130) & (0.8604-0.9078) & (0.8087-0.8596) \\
\bottomrule
    \end{tabular}%
\label{sup:tab10}
\end{table}

\begin{table}[htbp]
\centering
\scriptsize
\caption{Classification results of each endoscopist in the competent groups A and B on the held-out test set at the image level.}
\vspace{5pt}
    \begin{tabular}{cccccc}
    \toprule
    \multicolumn{6}{c}{\textbf{Competent group A w/o AI assistance}} \\
    \multicolumn{1}{c}{} & \textbf{Endoscopist 1} & \textbf{Endoscopist 2} & \textbf{Endoscopist 3} & \textbf{Endoscopist 4} & \textbf{Endoscopist 5} \\
    \midrule
    Accuracy & 0.7440 & 0.8107 & 0.7407 & 0.7319 & 0.6477 \\
    (95\% CI) & {(0.7157-0.7723)} & {(0.7853-0.8361) } & (0.7123-0.7691)  & (0.7032-0.7607)  & (0.6167-0.6787)  \\ \addlinespace[0.4em]
    Sensitivity & 0.7509 & 0.8293 & 0.7676 & 0.7423 & 0.7441 \\
    (95\% CI) & (0.7072-0.7945) & (0.7858-0.8728) & (0.7197-0.8155) & (0.6973-0.7872) & (0.7021-0.7861) \\ \addlinespace[0.4em]
    Specificity & 0.8747 & 0.9118 & 0.8779 & 0.8807 & 0.8460 \\
    (95\% CI) & (0.8481-0.9014) & (0.8912-0.9323) & (0.8529-0.903) & (0.8617-0.8998) & (0.8237-0.8682) \\
\bottomrule
    \end{tabular}%
\label{sup:tab11}
\end{table}

\begin{table}[htbp]
\centering
\scriptsize
\caption{Classification results of each endoscopist in the expert groups A and B on the held-out test set at the image level.}
\vspace{5pt}
    \begin{tabular}{cccccc}
    \toprule
    \multicolumn{6}{c}{\textbf{Expert group A w/o AI assistance}} \\
    \multicolumn{1}{c}{} & \textbf{Endoscopist 1} & \textbf{Endoscopist 2} & \textbf{Endoscopist 3} & \textbf{Endoscopist 4} & \textbf{Endoscopist 5} \\
    \midrule
    Accuracy & 0.8206 & 0.7954 & 0.7812 & 0.8545 & 0.8217 \\
    (95\% CI) & {(0.7957-0.8454)} & {(0.7693-0.8216)} & (0.7544-0.8080) & (0.8316-0.8773) & (0.7968-0.8465) \\ \addlinespace[0.4em]
    Sensitivity & 0.8323 & 0.8094 & 0.7729 & 0.8531 & 0.8479 \\
    (95\% CI) & (0.7906-0.8740) & (0.7691-0.8497) & (0.7247-0.8210) & (0.8136-0.8926) & (0.8087-0.8871) \\ \addlinespace[0.4em]
    Specificity & 0.9140 & 0.9025 & 0.8830 & 0.9241 & 0.9148 \\
    (95\% CI) & (0.8917-0.9364) & (0.8789-0.9260) & (0.8570-0.9089) & (0.9025-0.9457) & (0.8929-0.9368) \\
\bottomrule
    \end{tabular}%
\label{sup:tab12}
\end{table}

\begin{table*}[htbp]
\centering
\scriptsize
\caption{Comparison of our model with or without data standardisation (Ours v.s. Ours-Base). Evaluation is on the held-out test set at the image level across each hospital (Tongji, Ruijin, Huashan, Xinhua, Renji, and Zhongshan).}
\vspace{5pt}
    \begin{tabular}{cccccccccc}
    \toprule
    \multicolumn{1}{c}{\multirow{2}[0]{*}{\textbf{Hospital}}} & \multicolumn{1}{c}{\multirow{2}[0]{*}{\textbf{Model}}} & \textbf{Accuracy} & \textbf{Sensitivity} & \textbf{Specificity} & \textbf{PPV} & \textbf{NPV} & \multicolumn{1}{c}{\multirow{2}[0]{*}{\textbf{Kappa}}} & \multirow{2}[0]{*}{\textbf{AUC}} & \multicolumn{1}{c}{\multirow{2}[0]{*}{\textbf{AP}}} \\
          &       & (95\% CI) & (95\% CI) & (95\% CI) & (95\% CI) & (95\% CI) &       &       &  \\
    \midrule
    \multicolumn{1}{c}{\multirow{2}[2]{*}{\raisebox{-1em}{\textbf{All hospital}}}} & \multicolumn{1}{c}{\multirow{2}[1]{*}{\textbf{Ours-Base}}} & 0.8676 & 0.8546 & 0.9213 & 0.8636 & 0.9250 & \multirow{2}[1]{*}{0.7777} & \multirow{2}[1]{*}{0.9638} & \multirow{2}[1]{*}{0.9309} \\
    \multicolumn{1}{c}{\multirow{2}[2]{*}{\raisebox{-1em}{\textbf{(n=914)}}}} &       & (0.8456-0.8896) & (0.8159-0.8933) & (0.9008-0.9418) & (0.8247-0.9024) & (0.9046-0.9454) &       &       &  \\ \addlinespace[0.4em]
          & \multicolumn{1}{c}{\multirow{2}[1]{*}{\textbf{Ours}}} & 0.9256 & 0.9227 & 0.9581 & 0.9189 & 0.9570 & \multirow{2}[1]{*}{0.8763} & \multirow{2}[1]{*}{0.9818} & \multirow{2}[1]{*}{0.9675} \\
          &       & (0.9086-0.9426) & (0.8946-0.9509) & (0.9423-0.9738) & (0.8877-0.9501) & (0.9419-0.9721) &       &       &  \\
    \midrule
    \multicolumn{1}{c}{\multirow{2}[2]{*}{\raisebox{-1em}{\textbf{Tongji}}}} & \multicolumn{1}{c}{\multirow{2}[1]{*}{\textbf{Ours-Base}}} & 0.8905 & 0.8378 & 0.9192 & 0.8455 & 0.9205 & \multirow{2}[1]{*}{0.7682} & \multirow{2}[1]{*}{0.9617} & \multirow{2}[1]{*}{0.9234} \\
    \multicolumn{1}{c}{\multirow{2}[2]{*}{\raisebox{-1em}{\textbf{(n=201)}}}} &       & (0.8474-0.9337) & (0.7611-0.9146) & (0.8698-0.9687) & (0.7337-0.9572) & (0.8771-0.9639) &       &       &  \\
\bottomrule
    \end{tabular}%
\label{sup:tab13}
\end{table*}

\begin{table*}[htbp]
\setlength{\abovecaptionskip}{-20pt}
\centering
\tiny
\caption{Comparison of our model (Ours) and its two variants (Ours-v1 and Ours-v2) with or without data standardisation (Ours v.s. Ours-Base, Ours-v1 v.s. Ours-v1-Base, and Ours-v2 v.s. Ours-v2-Base). Evaluation is on the held-out test set at the patient level across each hospital (Tongji, Ruijin, Huashan, Xinhua, Renji, and Zhongshan).}
\vspace{5pt}
    \begin{tabular}{cccccccccc}
    \toprule
    \multirow{2}[0]{*}{\textbf{Hospital}} & \multicolumn{1}{c}{\multirow{2}[0]{*}{\textbf{Model}}} & \textbf{Accuracy} & \textbf{Sensitivity} & \textbf{Specificity} & \textbf{PPV} & \textbf{NPV} & \multicolumn{1}{c}{\multirow{2}[0]{*}{\textbf{Kappa}}} & \multirow{2}[0]{*}{\textbf{AUC}} & \multicolumn{1}{c}{\multirow{2}[0]{*}{\textbf{AP}}} \\
          &       & (95\% CI) & (95\% CI) & (95\% CI) & (95\% CI) & (95\% CI) &       &       &  \\
    \midrule
    \multicolumn{1}{c}{\multirow{12}{*}{\makecell{\textbf{All hospitals} \\ \textbf{(n=112)}}}} & \multicolumn{1}{c}{\multirow{2}[1]{*}{\textbf{Ours-Base}}} & 0.8929 & 0.8699 & 0.9520 & 0.8492 & 0.9454 & \multirow{2}[1]{*}{0.8303} & \multirow{2}[1]{*}{0.9830} & \multirow{2}[1]{*}{0.9668} \\
          &       & (0.8356-0.9501) & (0.7533-0.9843) & (0.9146-0.9895) & (0.7544-0.9439) & (0.8937-0.9955) &       &       &  \\
          & \multicolumn{1}{c}{\multirow{2}[0]{*}{\textbf{Ours}}} & 0.9464 & 0.9363 & 0.9719 & 0.9464 & 0.9728 & \multirow{2}[0]{*}{0.9136} & \multirow{2}[0]{*}{0.9948} & \multirow{2}[0]{*}{0.9905} \\
          &       & (0.9047-0.9881) & (0.8628-1.000) & (0.9444-0.9995) & (0.8808-1.000) & (0.9440-1.000) &       &       &  \\
          & \multicolumn{1}{c}{\multirow{2}[0]{*}{\textbf{Ours-v1-Base}}} & 0.8750 & 0.8639 & 0.9285 & 0.8537 & 0.9331 & \multirow{2}[0]{*}{0.8068} & \multirow{2}[0]{*}{0.9737} & \multirow{2}[0]{*}{0.9510} \\
          &       & (0.8137-0.9363) & (0.8181-0.9098) & (0.8948-0.9572) & (0.7831-0.9181) & (0.9059-0.9602) &       &       &  \\
\bottomrule
    \end{tabular}%
\label{sup:tab14}
\end{table*}

\begin{table}[p]
\centering
\scriptsize
\caption{Overall performance of our model on the WLI test set.}
\vspace{5pt}
    \begin{tabular}{ccccccccc}
    \toprule
    \multirow{2}[2]{*}{} & \textbf{Accuracy} & \textbf{Sensitivity} & \textbf{Specificity} & \textbf{PPV} & \textbf{NPV} & \multicolumn{1}{c}{\multirow{2}[2]{*}{\textbf{Kappa}}} & \multirow{2}[2]{*}{\textbf{AUC}} & \multicolumn{1}{c}{\multirow{2}[2]{*}{\textbf{AP}}} \\
          & (95\% CI) & (95\% CI) & (95\% CI) & (95\% CI) & (95\% CI) &       &       &  \\
    \midrule
    \multirow{2}[1]{*}{\textbf{Ours-WLI}} & 0.916 & 0.904 & 0.9509 & 0.9025 & 0.9502 & \multirow{2}[1]{*}{0.8563} & \multirow{2}[1]{*}{0.9842} & \multirow{2}[1]{*}{0.9702} \\
          & (0.8970-0.9351) & (0.8672-0.9409)) & (0.9340-0.9678) & (0.8699-0.9351) & (0.9320-0.9684) &       &       &  \\ \addlinespace[0.4em]
    \multirow{2}[1]{*}{\textbf{Ours}} & \textbf{0.9333} & \textbf{0.9174} & \textbf{0.9609} & \textbf{0.9212} & \textbf{0.962} & \multirow{2}[1]{*}{\textbf{0.8854}} & \multirow{2}[1]{*}{\textbf{0.9865}} & \multirow{2}[1]{*}{\textbf{0.9767}} \\
          & (0.9162-0.9505) & (0.8854-0.9495) & (0.9444-0.9773) & (0.8867-0.9557) & (0.9465-0.9774) &       &       &  \\
    \bottomrule
    \end{tabular}%
\label{sup:tab15}
\end{table}

\end{document}